# Efficient Markov Network Structure Discovery Using Independence Tests

**Facundo Bromberg**      FBROMBERG@FRM.UTN.EDU.AR
*Departamento de Sistemas de Información,*
*Universidad Tecnológica Nacional,*
*Mendoza, Argentina*

**Dimitris Margaritis**      DMARG@CS.IASTATE.EDU
**Vasant Honavar**      HONAVAR@CS.IASTATE.EDU
*Dept. of Computer Science,*
*Iowa State University,*
*Ames, IA 50011*

## Abstract

We present two algorithms for learning the structure of a Markov network from data: GSMN* and GSIMN. Both algorithms use statistical independence tests to infer the structure by successively constraining the set of structures consistent with the results of these tests. Until very recently, algorithms for structure learning were based on maximum likelihood estimation, which has been proved to be NP-hard for Markov networks due to the difficulty of estimating the parameters of the network, needed for the computation of the data likelihood. The independence-based approach does not require the computation of the likelihood, and thus both GSMN* and GSIMN can compute the structure efficiently (as shown in our experiments). GSMN* is an adaptation of the Grow-Shrink algorithm of Margaritis and Thrun for learning the structure of Bayesian networks. GSIMN extends GSMN* by additionally exploiting Pearl's well-known properties of the conditional independence relation to infer novel independences from known ones, thus avoiding the performance of statistical tests to estimate them. To accomplish this efficiently GSIMN uses the Triangle theorem, also introduced in this work, which is a simplified version of the set of Markov axioms. Experimental comparisons on artificial and real-world data sets show GSIMN can yield significant savings with respect to GSMN*, while generating a Markov network with comparable or in some cases improved quality. We also compare GSIMN to a forward-chaining implementation, called GSIMN-FCH, that produces all possible conditional independences resulting from repeatedly applying Pearl's theorems on the known conditional independence tests. The results of this comparison show that GSIMN, by the sole use of the Triangle theorem, is nearly optimal in terms of the set of independences tests that it infers.

## 1. Introduction

Graphical models (Bayesian and Markov networks) are an important subclass of statistical models that possess advantages that include clear semantics and a sound and widely accepted theoretical foundation (probability theory). Graphical models can be used to represent efficiently the joint probability distribution of a domain. They have been used in numerous application domains, ranging from discovering gene expression pathways in bioinformatics (Friedman, Linial, Nachman, & Pe'er, 2000) to computer vision (e.g. Geman

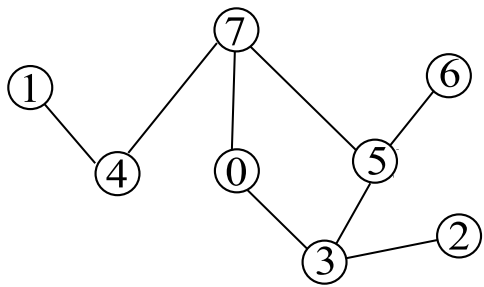

Figure 1: Example Markov network. The nodes represent variables in the domain $\mathbf{V} = \{0, 1, 2, 3, 4, 5, 6, 7\}$.

& Geman, 1984, Besag, York, & Mollie, 1991, Isard, 2003, Anguelov, Taskar, Chatalbashev, Koller, Gupta, Heitz, & Ng, 2005). One problem that naturally arises is the construction of such models from data (Heckerman, Geiger, & Chickering, 1995, Buntine, 1994). A solution to this problem, besides being theoretically interesting in itself, also holds the potential of advancing the state-of-the-art in application domains where such models are used.

In this paper we focus on the task of learning Markov networks (**MNs**) from data in domains in which all variables are either discrete or continuous and distributed according to a multidimensional Gaussian distribution. MNs are graphical models that consist of two parts: an undirected graph (the model structure), and a set of parameters. An example Markov network is shown in Figure 1. Learning such models from data consists of two interdependent tasks: learning the structure of the network, and, given the learned structure, learning the parameters. In this work we focus on the problem of learning the structure of the MN of a domain from data.

We present two algorithms for MN structure learning from data: **GSMN\*** (Grow-Shrink Markov Network learning algorithm) and **GSIMN** (Grow-Shrink Inference-based Markov Network learning algorithm). The GSMN\* algorithm is an adaptation to Markov networks of the GS algorithm by Margaritis and Thrun (2000), originally developed for learning the structure of Bayesian networks. GSMN\* works by first learning the local neighborhood of each variable in the domain (also called the *Markov blanket* of the variable), and then using this information in subsequent steps to improve efficiency. Although interesting and useful in itself, we use GSMN\* as a point of reference of the performance with regard to time complexity and accuracy achieved by GSIMN, which is the main result of this work. The GSIMN algorithm extends GSMN\* by using Pearl's theorems on the properties of the conditional independence relation (Pearl, 1988) to infer additional independences from a set of independences resulting from statistical tests and previous inferences, thus avoiding the execution of these tests on data. This allows savings in execution time and, when data are distributed, communication bandwidth.

The rest of the paper is organized as follows: In the next section we present previous research related to the problem. Section 3 introduces notation, definitions and presents some intuition behind the two algorithms. Section 4 contains the main algorithms, GSMN\* and GSIMN, as well as concepts and practical details related to their operation. We evaluate GSMN\* and GSIMN and present our results in Section 5, followed by a summary of our work and possible directions of future research in Section 6. Appendices A and B contain proofs of correctness of GSMN* and GSIMN.

## 2. Related Work

Markov networks have been used in the physics and computer vision communities (Geman & Geman, 1984, Besag et al., 1991, Anguelov et al., 2005) where they have been historically called Markov random fields. Recently there has been interest in their use for spatial data mining, which has applications in geography, transportation, agriculture, climatology, ecology and others (Shekhar, Zhang, Huang, & Vatsavai, 2004).

One broad and popular class of algorithms for learning the structure of graphical models is the *score-based* approach, exemplified for Markov networks by Della Pietra, Della Pietra, and Lafferty (1997), and McCallum (2003). Score-based approaches conduct a search in the space of legal structures in an attempt to discover a model structure of maximum score. Due to the intractable size of the search space i.e., the space of all legal graphs, which is super-exponential in size, score-based algorithms must usually resort to heuristic search. At each step of the structure search, a probabilistic inference step is necessary to evaluate the score (e.g., maximum likelihood, minimum description length, Lam & Bacchus, 1994, or pseudo-likelihood, Besag, 1974). For Bayesian networks this inference step is tractable and therefore several practical score-based algorithms for structure learning have been developed (Lam & Bacchus, 1994, Heckerman, 1995, Acid & de Campos, 2003). For Markov networks however, probabilistic inference requires the calculation of a normalizing constant (also known as partition function), a problem known to be NP-hard (Jerrum & Sinclair, 1993, Barahona, 1982). A number of approaches have considered a restricted class of graphical models (e.g. Chow & Liu, 1968, Rebane & Pearl, 1989, Srebro & Karger, 2001). However, Srebro and Karger (2001) prove that finding the maximum likelihood network is NP-hard for Markov networks of tree-width greater than 1.

Some work in the area of structure learning of undirected graphical models has concentrated on the learning of decomposable (also called chordal) MNs (Srebro & Karger, 2001). An example of learning non-decomposable MNs is presented in the work of Hofmann and Tresp (1998), which is an approach for learning structure in continuous domains with non-linear relationships among the domain attributes. Their algorithm removes edges greedily based on a leave-one-out cross validation log-likelihood score. A non-score based approach is in the work of Abbeel, Koller, and Ng (2006), which introduces a new class of efficient algorithms for structure and parameter learning of factor graphs, a class of graphical models that subsumes Markov and Bayesian networks. Their approach is based on a new parameterization of the Gibbs distribution in which the potential functions are forced to be probability distributions, and is supported by a generalization of the Hammersley-Clifford theorem for factor graphs. It is a promising and theoretically sound approach that may lead in the future to practical and efficient algorithms for undirected structure learning.

In this work we present algorithms that belong to the *independence-based* or *constraint-based* approach (Spirtes, Glymour, & Scheines, 2000). Independence-based algorithms exploit the fact that a graphical model implies that a set of independences exist in the distribution of the domain, and therefore in the data set provided as input to the algorithm (under assumptions, see next section); they work by conducting a set of conditional independence tests on data, successively restricting the number of possible structures consistent with the results of those tests to a singleton (if possible), and inferring that structure as the only possible one. A desirable characteristic of independence-based approaches is the fact that they do not require the use of probabilistic inference during the discovery of the structure. Also, such algorithms are amenable to proofs of correctness (under assumptions).

For Bayesian networks, the independence-based approach has been mainly exemplified by the SGS (Spirtes et al., 2000), PC (Spirtes et al., 2000), and algorithms that learn the Markov blanket as a step in learning the Bayesian network structure such as Grow-Shrink (GS) algorithm (Margaritis & Thrun, 2000), IAMB and its variants (Tsamardinos, Aliferis, & Statnikov, 2003a), HITON-PC and HITON-MB (Aliferis, Tsamardinos, & Statnikov, 2003), MMPC and MMMB (Tsamardinos, Aliferis, & Statnikov, 2003b), and max-min hill climbing (MMHC) (Tsamardinos, Brown, & Aliferis, 2006), all of which are widely used in the field. Algorithms for restricted classes such as trees (Chow & Liu, 1968) and polytrees (Rebane & Pearl, 1989) also exist.

For learning Markov networks previous work has mainly focused on learning *Gaussian graphical models*, where the assumption of a continuous multivariate Gaussian distribution is made; this results in linear dependences among the variables with Gaussian noise (Whittaker, 1990, Edwards, 2000). More recent approaches are included in the works of Dobra, Hans, Jones, Nevins, Yao, and West (2004), (Castelo & Roverato, 2006), Peña (2008), and Schäfer and Strimmer (2005), that focus on applications of Gaussian graphical models in Bioinformatics. While we do not make the assumption of continuous Gaussian variables in this paper, all algorithms we present are applicable to such domains with the use of an appropriate conditional independence test (such as partial correlation). The GSMN* and GSIMN algorithms presented apply to any case where an arbitrary faithful distribution can be assumed and a probabilistic conditional independence test for that distribution is available. The algorithms were first introduced by Bromberg, Margaritis, and Honavar (2006); the contributions of the present paper include extending these results by conducting an extensive evaluation of their experimental and theoretical properties. More specifically, the contributions include an extensive and systematic experimental evaluation of the proposed algorithms on (a) data sets sampled from artificially generated networks of varying complexity and strength of dependences, as well as (b) data sets sampled from networks representing real-world domains, and (c) formal proofs of correctness that guarantee that the proposed algorithms will compute the correct Markov network structure of the domain, under the stated assumptions.

## 3. Notation and Preliminaries

We denote random variables with capitals (e.g., $X, Y, Z$) and sets of variables with bold capitals (e.g., $\mathbf{X}, \mathbf{Y}, \mathbf{Z}$). In particular, we denote by $\mathbf{V} = \{0, \ldots, n-1\}$ the set of all $n$ variables in the domain. We name the variables by their indices in $\mathbf{V}$; for instance, we refer to the third variable in $\mathbf{V}$ simply by 3. We denote the data set as $D$ and its size (number of data points) by $|D|$ or $N$. We use the notation $(\mathbf{X} \perp\!\!\!\perp \mathbf{Y} \mid \mathbf{Z})$ to denote the proposition that $\mathbf{X}$ is independent of $\mathbf{Y}$ conditioned on $\mathbf{Z}$, for disjoint sets of variables $\mathbf{X}$, $\mathbf{Y}$, and $\mathbf{Z}$. $(\mathbf{X} \not\!\perp\!\!\!\perp \mathbf{Y} \mid \mathbf{Z})$ denotes conditional dependence. We use $(X \perp\!\!\!\perp Y \mid \mathbf{Z})$ as shorthand for $(\{X\} \perp\!\!\!\perp \{Y\} \mid \mathbf{Z})$ to improve readability.

A Markov network is an undirected graphical model that represents the joint probability distribution over $\mathbf{V}$. Each node in the graph represents one of the random variables in the domain, and absences of edges encode conditional independences among them. We assume the underlying probability distribution to be *graph-isomorph* (Pearl, 1988) or *faithful* (Spirtes et al., 2000), which means that it has a faithful undirected graph. A graph $G$ is said to be faithful to some distribution if its graph connectivity represents exactly those dependencies and independences existent in the distribution. In detail, this means that that for all disjoint sets $\mathbf{X}, \mathbf{Y}, \mathbf{Z} \subseteq \mathbf{V}$, $\mathbf{X}$ is independent of $\mathbf{Y}$ given $\mathbf{Z}$ if and only if the set of vertices $\mathbf{Z}$ separates the set of vertices $\mathbf{X}$ from the set of vertices $\mathbf{Y}$ in the graph $G$ (this is sometimes called the *global Markov property*, Lauritzen, 1996). In other words, this means that, after removing all vertices in $\mathbf{Z}$ from $G$ (including all edges incident to each of them), there exists no (undirected) path in the remaining graph between any variable in $\mathbf{X}$ to some variable in $\mathbf{Y}$. For example, in Figure 1, the set of variables $\{0, 5\}$ separates set $\{4, 6\}$ from set $\{2\}$. More generally, it has been shown (Pearl, 1988; Theorem 2, page 94 and definition of graph isomorphism, page 93) that a necessary and sufficient condition for a distribution to be graph-isomorph is for its set of independence relations to satisfy the following axioms for all disjoint sets of variables $\mathbf{X}$, $\mathbf{Y}$, $\mathbf{Z}$, $\mathbf{W}$ and individual variable $\gamma$:

$$
\begin{array}{llcl}
\textbf{(Symmetry)} & (\mathbf{X} \perp\!\!\!\perp \mathbf{Y} \mid \mathbf{Z}) & \Longleftrightarrow & (\mathbf{Y} \perp\!\!\!\perp \mathbf{X} \mid \mathbf{Z}) \\
\textbf{(Decomposition)} & (\mathbf{X} \perp\!\!\!\perp \mathbf{Y} \cup \mathbf{W} \mid \mathbf{Z}) & \Longleftrightarrow & (\mathbf{X} \perp\!\!\!\perp \mathbf{Y} \mid \mathbf{Z}) \,\wedge\, (\mathbf{X} \perp\!\!\!\perp \mathbf{W} \mid \mathbf{Z}) \\
\textbf{(Intersection)} & (\mathbf{X} \perp\!\!\!\perp \mathbf{Y} \mid \mathbf{Z} \cup \mathbf{W}) & & \\
& \wedge\, (\mathbf{X} \perp\!\!\!\perp \mathbf{W} \mid \mathbf{Z} \cup \mathbf{Y}) & \Longrightarrow & (\mathbf{X} \perp\!\!\!\perp \mathbf{Y} \cup \mathbf{W} \mid \mathbf{Z}) \\
\textbf{(Strong Union)} & (\mathbf{X} \perp\!\!\!\perp \mathbf{Y} \mid \mathbf{Z}) & \Longrightarrow & (\mathbf{X} \perp\!\!\!\perp \mathbf{Y} \mid \mathbf{Z} \cup \mathbf{W}) \\
\textbf{(Transitivity)} & (\mathbf{X} \perp\!\!\!\perp \mathbf{Y} \mid \mathbf{Z}) & \Longrightarrow & (\mathbf{X} \perp\!\!\!\perp \gamma \mid \mathbf{Z}) \vee (\gamma \perp\!\!\!\perp \mathbf{Y} \mid \mathbf{Z})
\end{array}
\tag{1}
$$

For the operation of the algorithms we also assume the existence of an oracle that can answer statistical independence queries. These are standard assumptions that are needed for formally proving the correctness of independence-based structure learning algorithms (Spirtes et al., 2000).

## 3.1 Independence-Based Approach to Structure Learning

GSMN* and GSIMN are independence-based algorithms for learning the structure of the Markov network of a domain. This approach works by evaluating a number of statistical independence statements, reducing the set of structures consistent with the results of these tests to a singleton (if possible), and inferring that structure as the only possible one.

As mentioned above, in theory we assume the existence of an independence-query oracle that can provide information about conditional independences among the domain variables. This can be viewed as an instance of a statistical query oracle (Kearns & Vazirani, 1994). In practice such an oracle does not exist; however, it can be implemented approximately by a statistical test evaluated on the data set $D$. For example, for discrete data this can be Pearson's conditional independence chi-square ($\chi^2$) test (Agresti, 2002), a mutual information test etc. For continuous Gaussian data a statistical test that can be used to measure conditional independence is partial correlation (Spirtes et al., 2000). To determine conditional independence between two variables $X$ and $Y$ given a set $\mathbf{Z}$ from data, the statistical test returns a p-value. The *p-value* of a test equals the probability of obtaining a value for the test statistic that is at least as extreme as the one that was actually observed given that the *null hypothesis* is true, which corresponds to conditional independence in our case. Assuming that the p-value of a test is $p(X, Y \mid \mathbf{Z})$, the statistical test concludes dependence if and only if $p(X, Y \mid \mathbf{Z})$ is less than or equal to a threshold $\alpha$ i.e.,

$$(X \not\!\perp Y \mid \mathbf{Z}) \iff p(X, Y \mid \mathbf{Z}) \leq \alpha.$$

The quantity $1 - \alpha$ is sometimes referred to as the test's *confidence threshold*. We use the standard value of $\alpha = 0.05$ in all our experiments, which corresponds to a confidence threshold of 95%.

In a faithful domain, it can be shown (Pearl & Paz, 1985) that an edge exists between two variables $X \neq Y \in \mathbf{V}$ in the Markov network of that domain if an only if they are dependent conditioned on all remaining variables in the domain, i.e.,

$$(X, Y) \text{ is an edge iff } (X \not\!\perp Y \mid \mathbf{V} - \{X, Y\}).$$

Thus, to learn the structure, theoretically it suffices to perform only $n(n-1)/2$ tests i.e., one test $(X, Y \mid \mathbf{V} - \{X, Y\})$ for each pair of variables $X, Y \in \mathbf{V}, X \neq Y$. Unfortunately, in non-trivial domains this usually involves a test that conditions on a large number of variables. Large conditioning sets produce sparse contingency tables (count histograms) which result in unreliable tests. This is because the number of possible configurations of the variables grows exponentially with the size of the conditioning set—for example, there are $2^n$ cells in a test involving $n$ binary variables, and to fill such a table with one data point per cell we would need a data set of at least exponential size i.e., $N \geq 2^n$. Exacerbating this problem, more than one data point per cell is typically necessary for a reliable test: As recommended by Cochran (1954), if more than 20% of the cells of the contingency table have less than 5 data points the test is deemed unreliable. Therefore both GSMN* and GSIMN algorithms (presented below) attempt to minimize the conditioning set size; they do that by choosing an order of examining the variables such that irrelevant variables are examined last.

## 4. Algorithms and Related Concepts

In this section we present our main algorithms, GSMN* and GSIMN, and supporting concepts required for their description. For the purpose of aiding the understanding of the reader, before discussing these we first describe the abstract GSMN algorithm in the next section. This helps in showing the intuition behind the algorithms and laying the foundation for them.

### 4.1 The Abstract GSMN Algorithm

For the sake of clarity of exposition, before discussing our first algorithm GSMN*, we describe the intuition behind it by describing its general structure using the abstract GSMN algorithm which deliberately leaves a number of details unspecified; these are filled-in in the concrete GSMN* algorithm, presented in the next section. Note that the choices for these

---

**Algorithm 1** GSMN algorithm outline: $G = GSMN(\mathbf{V}, D)$.

---

1: Initialize $G$ to the empty graph.
2: **for all** variables $X$ in the domain $\mathbf{V}$ **do**
3:      /* Learn the Markov Blanket $\mathbf{B}^X$ of $X$ using the GS algorithm. */
4:      $\mathbf{B}^X \leftarrow GS(X, \mathbf{V}, D)$
5:      Add an undirected edge in $G$ between $X$ and each variable $Y \in \mathbf{B}^X$.
6: **return** $G$

---

**Algorithm 2** GS algorithm. Returns the Markov Blanket $\mathbf{B}^X$ of variable $X \in \mathbf{V}$: $\mathbf{B}^X = GS(X, \mathbf{V}, D)$.

---

1: $\mathbf{B}^X \leftarrow \varnothing$
2: /* Grow phase. */
3: **for** each variable $Y$ in $\mathbf{V} - \{X\}$ **do**
4:      **if** $(X \not\perp Y \mid \mathbf{B}^X)$ (estimated using data $D$) **then**
5:          $\mathbf{B}^X \leftarrow \mathbf{B}^X \cup \{Y\}$
6:          **goto** 3 /* Restart grow loop. */
7: /* Shrink phase. */
8: **for** each variable $Y$ in $\mathbf{B}^X$ **do**
9:      **if** $(X \perp\!\!\!\perp Y \mid \mathbf{B}^X - \{Y\})$ (estimated using data $D$) **then**
10:         $\mathbf{B}^X \leftarrow \mathbf{B}^X - \{Y\}$
11:         **goto** 8 /* Restart shrink loop. */
12: **return** $\mathbf{B}^X$

---

details are a source of optimizations that can reduce the algorithm's computational cost. We make these explicit when we discuss the concrete GSMN* and GSIMN algorithms.

The abstract GSMN algorithm is shown in Algorithm 1. Given as input a data set $D$ and a set of variables $\mathbf{V}$, GSMN computes the set of nodes (variables) $\mathbf{B}^X$ that are adjacent to each variable $X \in \mathbf{V}$; these completely determine the structure of the domain MN. The algorithm consists of a main loop in which it learns the Markov blanket $\mathbf{B}^X$ of each node (variable) $X$ in the domain using the GS algorithm. It then constructs the Markov network structure by connecting $X$ with each variable in $\mathbf{B}^X$.

The GS algorithm was first proposed by Margaritis and Thrun (2000) and is shown in Algorithm 2. It consists of two phases, a *grow phase* and a *shrink phase*. The grow phase of $X$ proceeds by attempting to add each variable $Y$ to the current set of hypothesized neighbors of $X$, contained in $\mathbf{B}^X$, which is initially empty. $\mathbf{B}^X$ grows by some variable $Y$ during each iteration of the grow loop of $X$ if and only if $Y$ is found dependent with $X$ given the current set of hypothesized neighbors $\mathbf{B}^X$. Due to the (unspecified) ordering that the variables are examined (this is explicitly specified in the concrete GSMN* algorithm, presented in the next section), at the end of the grow phase some of the variables in $\mathbf{B}^X$ might not be true neighbors of $X$ in the underlying MN—these are called *false positives*. This justifies the shrink phase of the algorithm, which removes each false positive $Y$ in $\mathbf{B}^X$ by testing for independence with $X$ conditioned on $\mathbf{B}^X - \{Y\}$. If $Y$ is found independent of $X$ during the shrink phase, it cannot be a true neighbor (i.e., there cannot be an edge $X-Y$), and GSMN removes it from $\mathbf{B}^X$. Assuming faithfulness and correctness of the independence query results, by the end of the shrink phase $\mathbf{B}^X$ contains exactly the neighbors of $X$ in the underlying Markov network.

In the next section we present a concrete implementation of GSMN, called GSMN*. This augments GSMN by specifying a concrete ordering that the variables $X$ are examined in the main loop of GSMN (lines 2–5 in Algorithm 1), as well as a concrete order that the variables $Y$ are examined in the grow and shrink phases of the GS algorithm (lines 3–6 and 8–11 in Algorithm 2, respectively).

## 4.2 The Concrete GSMN* Algorithm

In this section we discuss our first algorithm, GSMN* (Grow-Shrink Markov Network learning algorithm), for learning the structure of the Markov network of a domain. Note that the reason for introducing GSMN* in addition to our main contribution, the GSIMN algorithm (presented later in Section 4.5), is for comparison reasons. In particular, GSIMN and GSMN* have identical structure, following the same order of examination of variables, with their only difference being the use of inference by GSIMN (see details in subsequent sections). Introducing GSMN* therefore makes it possible to measure precisely (through our experimental results in Section 5) the benefits of the use of inference on performance.

The GSMN* algorithm is shown in Algorithm 3. Its structure is similar to the abstract GSMN algorithm. One notable difference is that the order that variables are examined is now specified; this is done in the initialization phase where the so-called *examination order* $\pi$ and *grow order* $\lambda_X$ of each variable $X \in \mathbf{V}$ is determined. $\pi$ and all $\lambda_X$ are priority queues and each is initially a permutation of $\mathbf{V}$ ($\lambda_X$ is a permutation of $\mathbf{V} - \{X\}$) such that the position of a variable in the queue denotes its priority e.g., $\pi = [2, 0, 1]$ means that variable 2 has the highest priority (will be examined first), followed by 0 and finally by 1. Similarly, the position of a variable in $\lambda_X$ determines the order it will be examined during the grow phase of $X$.

During the initialization phase the algorithm computes the strength of unconditional dependence between each pair of variable $X$ and $Y$, as given by the unconditional p-value $p(X, Y \mid \varnothing)$ of an independence test between each pair of variables $X \neq Y$, denoted by $p_{XY}$ in the algorithm. (In practice the logarithm of the p-values is computed, which allows greater precision in domains where some dependencies may be very strong or very weak.) In particular, the algorithm gives higher priority to (examines earlier) those variables with a lower average log p-value (line 5), indicating stronger dependence. This average is defined as:

$$\operatorname*{avg}_{Y} \log(p_{XY}) = \frac{1}{|\mathbf{V}| - 1} \sum_{Y \neq X} \log(p_{XY}).$$

For the grow order $\lambda_X$ of variable $X$, the algorithm gives higher priority to those variables $Y$ whose p-value (or equivalently the log of the p-value) with variable $X$ is small (line 8). This ordering is due to the intuition behind the "folk-theorem" (as Koller & Sahami, 1996, puts it) that states that probabilistic influence or association between attributes tends to attenuate over distance in a graphical model. This suggests that a pair of variables $X$ and $Y$ with high unconditional p-value are less likely to be directly linked. Note that this ordering is a heuristic and is not guaranteed to hold in general. For example, it may not hold if the underlying domain is a Bayesian network e.g., two "spouses" may be independent unconditionally but dependent conditional on a common child. Note however that this example does not apply to faithful domains i.e., graph-isomorph to a Markov network. Also

---

**Algorithm 3** GSMN\*, a concrete implementation of GSMN: $G = GSMN^*(\mathbf{V}, D)$.

---

1: Initialize $G$ to the empty graph.
2: /\* *Initialization.* \*/
3: **for all** $X, Y \in \mathbf{V}, X \neq Y$ **do**
4:      $p_{XY} \leftarrow p(X, Y \mid \varnothing)$
5: Initialize $\pi$ such that $\forall i, i' \in \{0, \ldots, n-1\}, \left[i < i' \iff \underset{j}{\operatorname{avg}} \log(p_{\pi_i j}) < \underset{j}{\operatorname{avg}} \log(p_{\pi_{i'} j})\right]$.

6: **for all** $X \in \mathbf{V}$ **do**
7:      $\mathbf{B}^X \leftarrow \varnothing$
8:      Initialize $\lambda^X$ such that $\forall j, j' \in \{0, \ldots, n-1\}, \left[j < j' \iff p_{X\lambda_j^X} < p_{X\lambda_{j'}^X}\right]$.
9:      Remove $X$ from $\lambda^X$.
10: /\* *Main loop.* \*/
11: **while** $\pi$ is not empty **do**
12:      $X \leftarrow dequeue(\pi)$
13:      /\* *Propagation phase.* \*/
14:      $\mathbf{T} \leftarrow \{Y : Y \text{ was examined and } X \in \mathbf{B}^Y\}$
15:      $\mathbf{F} \leftarrow \{Y : Y \text{ was examined and } X \notin \mathbf{B}^Y\}$
16:      **for all** $Y \in \mathbf{T}$, move $Y$ to the end of $\lambda^X$.
17:      **for all** $Y \in \mathbf{F}$, move $Y$ to the end of $\lambda^X$.
18:      /\* *Grow phase.* \*/
19:      $\mathbf{S} \leftarrow \varnothing$
20:      **while** $\lambda^X$ not empty **do**
21:          $Y \leftarrow dequeue(\lambda^X)$
22:          **if** $p_{XY} \leq \alpha$ **then**
23:              **if** $\neg I_{\text{GSMN}^*}(X, Y, \mathbf{S}, \mathbf{F}, \mathbf{T})$ **then**
24:                  $\mathbf{S} \leftarrow \mathbf{S} \cup \{Y\}$
25:                  /\* *Change grow order of $Y$.* \*/
26:                  Move $X$ to the beginning of $\lambda_Y$.
27:                  **for** $W = S_{|\mathbf{S}|-2}$ to $S_0$ **do**
28:                      Move $W$ to the beginning of $\lambda_Y$.
29:          /\* *Change examination order.* \*/
30:          **for** $W = S_{|\mathbf{S}|-1}$ to $S_0$ **do**
31:              **if** $W \in \pi$ **then**
32:                  Move $W$ to the beginning of $\pi$.
33:                  **break** to line 34
34:      /\* *Shrink phase.* \*/
35:      **for** $Y = S_{|\mathbf{S}|-1}$ to $S_0$ **do**
36:          **if** $I_{\text{GSMN}^*}(X, Y, \mathbf{S} - \{Y\}, \mathbf{F}, \mathbf{T})$ **then**
37:              $\mathbf{S} \leftarrow \mathbf{S} - \{Y\}$
38:      $\mathbf{B}^X \leftarrow \mathbf{S}$
39:      Add an undirected edge in $G$ between $X$ and each variable $Y \in \mathbf{B}^X$.
40: **return** $G$

---

note that the correctness of all algorithms we present does not depend on it holding i.e., as we prove in Appendices A and B, both GSMN\* and GSIMN are guaranteed to return the correct structure under the assumptions stated in Section 3 above. Also note that the computational cost for the calculation of $p_{XY}$ is low due to the empty conditioning set.

The remaining of the GSMN\* algorithm contains the main loop (lines 10–39) in which each variable in $\mathbf{V}$ is examined according to the examination order $\pi$, determined during

---

**Algorithm 4** $I_{\text{GSMN}^*}(X, Y, \mathbf{S}, \mathbf{F}, \mathbf{T})$: Calculate independence test $(X, Y \mid \mathbf{S})$ by propagation, if possible, otherwise run a statistical test on data.

---

1: /* *Attempt to infer dependence by propagation.* */
2: **if** $Y \in \mathbf{T}$ **then**
3:     return false
4: /* *Attempt to infer independence by propagation.* */
5: **if** $Y \in \mathbf{F}$ **then**
6:     return true
7: /* *Else do statistical test on data.* */
8: $t \leftarrow \mathbf{1}_{(p(X,Y|\mathbf{Z}) > \alpha)}$ /* $t = $ true *iff p-value of statistical test* $(X, Y \mid \mathbf{S}) > \alpha$. */
9: **return** $t$

---

the initialization phase. The main loop includes three phases: the *propagation phase* (lines 13–17), the *grow phase* (lines 18–33), and the *shrink phase* (lines 34–37). The propagation phase is an optimization in which all variables $Y$ for which $\mathbf{B}^Y$ has already been computed (i.e., all variables $Y$ already examined) are collected in two sets $\mathbf{F}$ and $\mathbf{T}$. Set $\mathbf{F}$ ($\mathbf{T}$) contains all variables $Y$ such that $X \notin \mathbf{B}^Y$ ($X \in \mathbf{B}^Y$). Both sets are passed to the independence procedure $I_{\text{GSMN}^*}$, shown in Algorithm 4, for the purpose of avoiding the execution of any tests between $X$ and $Y$ by the algorithm. This is justified by the fact that, in undirected graphs, $Y$ is in the Markov blanket of $X$ if and only if $X$ is in the Markov blanket of $Y$. Variables $Y$ already found not to contain $X$ in their blanket $\mathbf{B}^Y$ (set $\mathbf{F}$) cannot be members of $\mathbf{B}^X$ because there exists some set of variables that has rendered them conditionally independent of $X$ in a previous step, and independence can therefore be inferred easily. Note that in the experiments section of the paper (Section 5) we evaluate GSMN* with and without the propagation phase, in order to measure the effect that this propagation optimization has on performance. Turning off propagation is accomplished simply by setting sets $\mathbf{T}$ and $\mathbf{F}$ (as computed in lines 14 and 15, respectively) to the empty set.

Another difference of GSMN* from the abstract GSMN algorithm is in the use of condition $p_{XY} \leq \alpha$ (line 22). This is an additional optimization that avoids an independence test in the case that $X$ and $Y$ were found (unconditionally) independent during the initialization phase, since in that case this would imply $X$ and $Y$ are independent given any conditioning set by the axiom of Strong Union.

A crucial difference between GSMN* and the abstract GSMN algorithm is that GSMN* changes the examination order $\pi$ and the grow order $\lambda_Y$ of every variable $Y \in \lambda_X$. (Since $X \notin \lambda_X$, this excludes the grow order of $X$ itself.) These changes in ordering proceed as follows: After the end of the grow phase of variable $X$, the new examination order $\pi$ (set in lines 30–33) dictates that the next variable $W$ to be examined after $X$ is the last to be added to $\mathbf{S}$ during the growing phase that has not yet been examined (i.e., $W$ is still in $\pi$). The grow order $\lambda_Y$ of all variables $Y$ found dependent with $X$ is also changed; this is done to maximize the number of optimizations by the GSIMN algorithm (our main contribution in this paper) which shares the algorithm structure of GSMN*. The changes in grow order are therefore explained in detail in Section 4.5 when GSIMN is presented.

A final difference between GSMN* and the abstract GSMN algorithm is the restart actions of the grow and shrink phases of GSMN whenever the current Markov blanket is modified (lines 6 and 11 of Algorithm 2), which are not present in GSMN*. The restarting

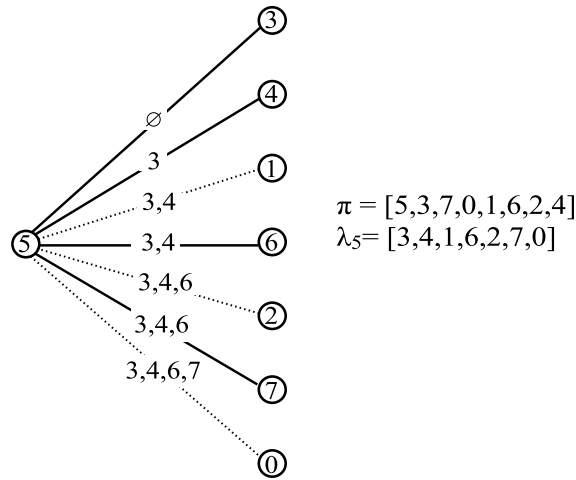

$\pi = [5,3,7,0,1,6,2,4]$
$\lambda_5 = [3,4,1,6,2,7,0]$

Figure 2: Illustration of the operation of GSMN* using an independence graph. The figure shows the growing phase of variable 5. Variables are examined according to its grow order $\lambda_5 = [3, 4, 1, 6, 2, 7, 0]$.

of the loops was necessary in the GS algorithm due to its original usage in learning the structure of Bayesian networks. In that task, it was possible for a true member $Y$ of the blanket of $X$ to be found initially independent during the grow loop when conditioning on some set $\mathbf{S}$ but to be found dependent later when conditioned on a superset $\mathbf{S}' \supset \mathbf{S}$. This could happen if $Y$ was an "unshielded spouse" of $X$ i.e., if $Y$ had one or more common children with $X$ but there existed no direct link between $Y$ and $X$ in the underlying Bayesian network. However, this behavior is impossible in a domain that has a distribution faithful to a Markov network (one of our assumptions): any independence between $X$ and $Y$ given $\mathbf{S}$ must hold for any superset $\mathbf{S}'$ of $\mathbf{S}$ by the axiom of Strong Union (see Eqs. (1)). The restart of the grow and shrink loops is therefore omitted from GSMN* in order to save unnecessary tests. Note that, even though it is possible that this behavior is impossible in faithful domains, it is possible in unfaithful ones, so we also experimentally evaluated our algorithms in real-world domains in which the assumption of Markov faithfulness may not necessarily hold (Section 5).

A proof of correctness of GSMN* is presented in Appendix A.

## 4.3 Independence Graphs

We can demonstrate the operation of GSMN* graphically by the concept of the *independence graph*, which we now introduce. We define an independence graph to be an undirected graph in which conditional independences and dependencies between single variables are represented by one or more *annotated edges* between them. A solid (dotted) edge between variables $X$ and $Y$ annotated by $\mathbf{Z}$ represents the fact that $X$ and $Y$ have been found dependent (independent) given $\mathbf{Z}$. If the conditioning set $\mathbf{Z}$ is enclosed in parentheses then this edge represents an independence or dependence that was *inferred* from Eqs. (1) (as opposed to computed from statistical tests). Shown graphically:

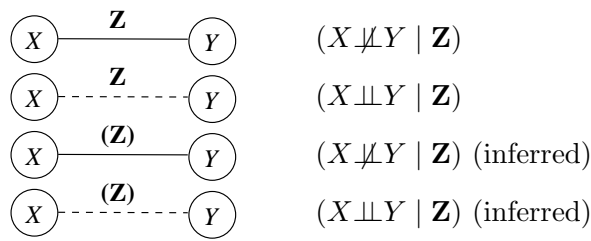

For instance, in Figure 2, the dotted edge between 5 and 1 annotated with "3, 4" represents the fact that $(5 \perp\!\!\!\perp 1 \mid \{3, 4\})$. The absence of an edge between two variables indicates the absence of information about the independence or dependence between these variables under any conditioning set.

**Example 1.** *Figure 2 illustrates the operation of GSMN\* using an independence graph in the domain whose underlying Markov network is shown in Figure 1. The figure shows the independence graph at the end of the grow phase of the variable 5, the first in the examination order $\pi$. (We do not discuss in this example the initialization phase of GSMN\*; instead, we assume that the examination ($\pi$) and grow ($\lambda$) orders are as shown in the figure.) According to vertex separation on the underlying network (Figure 1), variables 3, 4, 6, and 7 are found dependent with 5 during the growing phase i.e.,*

$$\neg I(5, 3 \mid \varnothing),$$
$$\neg I(5, 4 \mid \{3\}),$$
$$\neg I(5, 6 \mid \{3, 4\}),$$
$$\neg I(5, 7 \mid \{3, 4, 6\})$$

*and are therefore connected to 5 in the independence graph by solid edges annotated by sets $\varnothing$, $\{3\}$, $\{3, 4\}$ and $\{3, 4, 6\}$ respectively. Variables 1, 2, and 0 are found independent i.e.,*

$$I(5, 1 \mid \{3, 4\}),$$
$$I(5, 2 \mid \{3, 4, 6\}),$$
$$I(5, 0 \mid \{3, 4, 6, 7\})$$

*and are thus connected to 5 by dotted edges annotated by $\{3, 4\}$, $\{3, 4, 6\}$ and $\{3, 4, 6, 7\}$ respectively.*

## 4.4 The Triangle Theorem

In this section we present and prove a theorem that is used in the subsequent GSIMN algorithm. As will be seen, the main idea behind the GSIMN algorithm is to attempt to decrease the number of tests done by exploiting the properties of the conditional independence relation in faithful domains i.e., Eqs. (1). These properties can be seen as inference rules that can be used to derive new independences from ones that we know to be true. A careful study of these axioms suggests that only two simple inference rules, stated in the *Triangle theorem* below, are sufficient for inferring most of the useful independence information that can be inferred by a systematic application of the inference rules. This is confirmed in our experiments in Section 5.

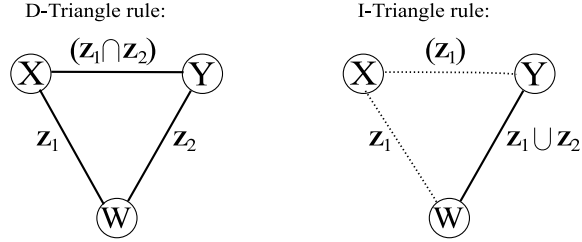

Figure 3: Independence graph depicting the Triangle theorem. Edges in the graph are labeled by sets and represent conditional independences or dependencies. A solid (dotted) edge between $X$ and $Y$ labeled by $\mathbf{Z}$ means that $X$ and $Y$ are dependent (independent) given $\mathbf{Z}$. A set label enclosed in parentheses means the edge was inferred by the theorem.

**Theorem 1** (Triangle theorem). *Given Eqs. (1), for every variable $X$, $Y$, $W$ and sets $\mathbf{Z}_1$ and $\mathbf{Z}_2$ such that $\{X, Y, W\} \cap \mathbf{Z}_1 = \{X, Y, W\} \cap \mathbf{Z}_2 = \varnothing$,*

$$(X \not\perp\!\!\!\perp W \mid \mathbf{Z}_1) \wedge (W \not\perp\!\!\!\perp Y \mid \mathbf{Z}_2) \implies (X \not\perp\!\!\!\perp Y \mid \mathbf{Z}_1 \cap \mathbf{Z}_2)$$
$$(X \perp\!\!\!\perp W \mid \mathbf{Z}_1) \wedge (W \not\perp\!\!\!\perp Y \mid \mathbf{Z}_1 \cup \mathbf{Z}_2) \implies (X \perp\!\!\!\perp Y \mid \mathbf{Z}_1).$$

*We call the first relation the "D-triangle rule" and the second the "I-triangle rule."*

*Proof.* We are using the Strong Union and Transitivity of Eqs. (1) as shown or in contrapositive form.

**(Proof of D-triangle rule)**:

- From Strong Union and $(X \not\perp\!\!\!\perp W \mid \mathbf{Z}_1)$ we get $(X \not\perp\!\!\!\perp W \mid \mathbf{Z}_1 \cap \mathbf{Z}_2)$.
- From Strong Union and $(W \not\perp\!\!\!\perp Y \mid \mathbf{Z}_1)$ we get $(W \not\perp\!\!\!\perp Y \mid \mathbf{Z}_1 \cap \mathbf{Z}_2)$.
- From Transitivity, $(X \not\perp\!\!\!\perp W \mid \mathbf{Z}_1 \cap \mathbf{Z}_2)$, and $(W \not\perp\!\!\!\perp Y \mid \mathbf{Z}_1 \cap \mathbf{Z}_2)$, we get $(X \not\perp\!\!\!\perp Y \mid \mathbf{Z}_1 \cap \mathbf{Z}_2)$.

**(Proof of I-triangle rule)**:

- From Strong Union and $(W \not\perp\!\!\!\perp Y \mid \mathbf{Z}_1 \cup \mathbf{Z}_2)$ we get $(W \not\perp\!\!\!\perp Y \mid \mathbf{Z}_1)$.
- From Transitivity, $(X \perp\!\!\!\perp W \mid \mathbf{Z}_1)$ and $(W \not\perp\!\!\!\perp Y \mid \mathbf{Z}_1)$ we get $(X \perp\!\!\!\perp Y \mid \mathbf{Z}_1)$.

$\square$

We can represent the Triangle theorem graphically using the independence graph construct of Section 4.2. Figure 3 depicts the two rules of the Triangle theorem using two independence graphs.

The Triangle theorem can be used to infer additional conditional independences from tests conducted during the operation of GSMN*. An example of this is shown in Figure 4, which illustrates the application of the Triangle theorem to the example presented in Figure 2. The independence information inferred from the Triangle theorem is shown by curved edges (note that the conditioning set of each such edge is enclosed in parentheses).

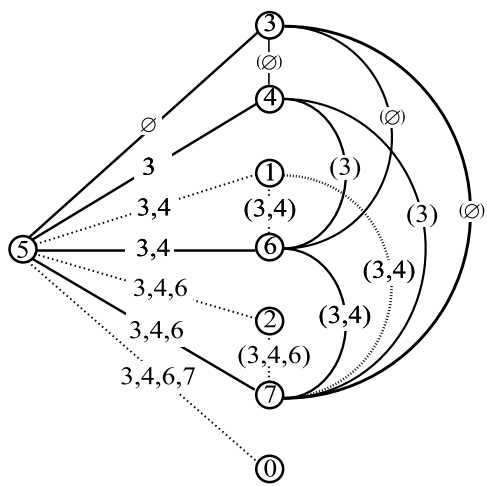

Figure 4: Illustration of the use of the Triangle theorem on the example of Figure 2. The set of variables enclosed in parentheses correspond to tests inferred by the Triangle theorem using the two adjacent edges as antecedents. For example, the result $(1 \perp\!\!\!\perp 7 \mid \{3, 4\})$, is inferred from the I-triangle rule, independence $(5 \perp\!\!\!\perp 1 \mid \{3, 4\})$ and dependence $(5 \not\!\perp\!\!\!\perp 7 \mid \{3, 4, 6\})$.

For example, independence edge $(4, 7)$ can be inferred by the D-triangle rule from the adjacent edges $(5, 4)$ and $(5, 7)$, annotated by $\{3\}$ and $\{3, 4, 6\}$ respectively. The annotation for this inferred edge is $\{3\}$, which is the intersection of the annotations $\{3\}$ and $\{3, 4, 6\}$. An example application of the I-triangle rule is edge $(1, 7)$, which is inferred from edges $(5, 1)$ and $(5, 7)$ with annotations $\{3, 4\}$ and $\{3, 4, 6\}$ respectively. The annotation for this inferred edge is $\{3, 4\}$, which is the intersection of the annotations $\{3, 4, 6\}$ and $\{3, 4\}$.

## 4.5 The GSIMN Algorithm

In the previous section we saw the possibility of using the two rules of the Triangle theorem to infer the result of novel tests during the grow phase. The GSIMN algorithm (Grow-Shrink Inference-based Markov Network learning algorithm), introduced in this section, uses the Triangle theorem in a similar fashion to extend GSMN* by inferring the value of a number of tests that GSMN* executes, making their evaluation unnecessary. GSIMN and GSMN* work in exactly the same way (and thus the GSIMN algorithm shares exactly the same algorithmic description i.e., both follow Algorithm 3), with all differences between them concentrated in the independence procedure they use: instead of using independence procedure $I_{\text{GSMN*}}$ of GSMN*, GSIMN uses procedure $I_{\text{GSIMN}}$, shown in Algorithm 5. Procedure $I_{\text{GSIMN}}$, in addition to attempting to propagate the blanket information obtained from the examination of previous variables (as $I_{\text{GSMN*}}$ does), also attempts to infer the value of the independence test that is provided as its input by either the Strong Union axiom (listed in Eqs. (1)) or the Triangle theorem. If this attempt is successful, $I_{\text{GSIMN}}$ returns the value inferred (`true` or `false`), otherwise it defaults to a statistical test on the data set (as $I_{\text{GSMN*}}$ does). For the purpose of assisting in the inference process, GSIMN and

**Algorithm 5** $I_{\text{GSIMN}}(X, Y, \mathbf{S}, \mathbf{F}, \mathbf{T})$: Calculate independence test result by inference (including propagation), if possible. Record test result in the knowledge base.

---

1: /* *Attempt to infer dependence by propagation.* */
2: **if** $Y \in \mathbf{T}$ **then**
3:   **return** `false`
4: /* *Attempt to infer independence by propagation.* */
5: **if** $Y \in \mathbf{F}$ **then**
6:   **return** `true`
7: /* *Attempt to infer dependence by Strong Union.* */
8: **if** $\exists (\mathbf{A}, \texttt{false}) \in K_{XY}$ such that $\mathbf{A} \supseteq \mathbf{S}$ **then**
9:   **return** `false`
10: /* *Attempt to infer dependence by the D-triangle rule.* */
11: **for all** $W \in \mathbf{S}$ **do**
12:   **if** $\exists (\mathbf{A}, \texttt{false}) \in K_{XW}$ such that $\mathbf{A} \supseteq \mathbf{S} \;\wedge\; \exists (\mathbf{B}, \texttt{false}) \in K_{WY}$ such that $\mathbf{B} \supseteq \mathbf{S}$ **then**
13:     Add $(\mathbf{A} \cap \mathbf{B}, \texttt{false})$ to $K_{XY}$ and $K_{YX}$.
14:     **return** `false`
15: /* *Attempt to infer independence by Strong Union.* */
16: **if** $\exists (\mathbf{A}, \texttt{true}) \in K_{XY}$ such that $\mathbf{A} \subseteq \mathbf{S}$ **then**
17:   **return** `true`
18: /* *Attempt to infer independence by the I-triangle rule.* */
19: **for all** $W \in \mathbf{S}$ **do**
20:   **if** $\exists (\mathbf{A}, \texttt{true}) \in K_{XW}$ s.t. $\mathbf{A} \subseteq \mathbf{S} \;\wedge\; \exists (\mathbf{B}, \texttt{false}) \in K_{WY}$ s.t. $\mathbf{B} \supseteq \mathbf{A}$ **then**
21:     Add $(\mathbf{A}, \texttt{true})$ to $K_{XY}$ and $K_{YX}$.
22:     **return** `true`
23: /* *Else do statistical test on data.* */
24: $t \leftarrow \mathbf{1}_{(p(X,Y|\mathbf{Z}) > \alpha)}$ /* *$t = \texttt{true}$ iff p-value of statistical test $(X, Y \mid \mathbf{S}) > \alpha$.* */
25: Add $(\mathbf{S}, t)$ to $K_{XY}$ and $K_{YX}$.
26: **return** $t$

---

$I_{\text{GSIMN}}$ maintain a knowledge base $K_{XY}$ for each pair of variables $X$ and $Y$, containing the outcomes of all tests evaluated so far between $X$ and $Y$ (either from data or inferred). Each of these knowledge bases is empty at the beginning of the GSIMN algorithm (the initialization step is not shown in the algorithm since GSMN* does not use it), and is maintained within the test procedure $I_{\text{GSIMN}}$.

We now explain $I_{\text{GSIMN}}$ (Algorithm 5) in detail. $I_{\text{GSIMN}}$ attempts to infer the independence value of its input triplet $(X, Y \mid \mathbf{S})$ by applying a single step of backward chaining using the Strong Union and Triangle rules i.e., it searches the knowledge base $K = \{K_{XY} : X, Y \in \mathbf{V}\}$ for antecedents of instances of rules that have the input triplet $(X, Y \mid \mathbf{S})$ as consequent. The Strong Union rule is used in its direct form as shown in Eqs. (1) and also in its contrapositive form. The direct form can be used to infer independences, and therefore we refer to it as the I-SU rule from here on. In its contrapositive form, the I-SU rule becomes $(X \not\!\perp Y \mid \mathbf{S} \cup \mathbf{W}) \implies (X \not\!\perp Y \mid \mathbf{S})$, referred to as the D-SU rule since it can be used to infer dependencies. According to the D-Triangle and D-SU rules, the dependence $(X \not\!\perp Y \mid \mathbf{S})$ can be inferred if the knowledge base $K$ contains

1. a test $(X \not\!\perp Y \mid \mathbf{A})$ with $\mathbf{A} \supseteq \mathbf{S}$, or

2. tests $(X \not\!\perp W \mid \mathbf{A})$ and $(W \not\!\perp Y \mid \mathbf{B})$ for some variable $W$, with $\mathbf{A} \supseteq \mathbf{S}$ and $\mathbf{B} \supseteq \mathbf{S}$,

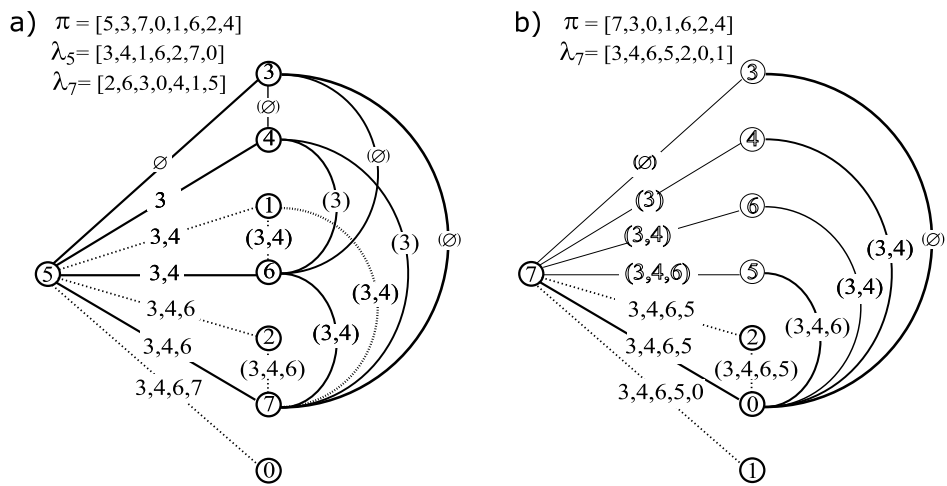

Figure 5: Illustration of the operation of GSIMN. The figure shows the grow phase of two consecutively examined variables 5 and 7. The figure shows how the variable examined second is not 3 but 7, according to the change in the examination order $\pi$ in lines 30–33 of Algorithm 3. The set of variables enclosed in parentheses correspond to tests inferred by the Triangle theorem using two adjacent edges as antecedents. The results $(7 \!\not\perp\!\! 3 \mid \varnothing)$, $(7 \!\not\perp\!\! 4 \mid \{3\})$, $(7 \!\not\perp\!\! 6 \mid \{3,4\})$, and $(7 \!\not\perp\!\! 5 \mid \{3,4,6\})$ in (b), shown highlighted, were not executed but inferred from the tests done in (a).

respectively. According to the I-Triangle and I-SU rules, the independence $(X \!\perp\!\! Y \mid \mathbf{S})$ can be inferred if the knowledge base contains

3. a test $(X \!\perp\!\! Y \mid \mathbf{A})$ with $\mathbf{A} \subseteq \mathbf{S}$, or

4. tests $(X \!\perp\!\! W \mid \mathbf{A})$ and $(W \!\not\perp\!\! Y \mid \mathbf{B})$ for some variable $W$, with $\mathbf{A} \subseteq \mathbf{S}$ and $\mathbf{B} \supseteq \mathbf{A}$, respectively.

The changes to the grow orders of some variables occur inside the grow phase of the currently examined variable $X$ (lines 25–28 of GSIMN i.e., Algorithm 3 with $I_{\text{GSIMN}^*}$ replaced by $I_{\text{GSIMN}}$). In particular, if, for some variable $Y$, the algorithm reaches line 24, i.e., $p_{XY} \leq \alpha$ and $I_{\text{GSIMN}}(X, Y, \mathbf{S}) = \texttt{false}$, then $X$ and all the variables that were found dependent with $X$ before $Y$ (i.e., all variables currently in $\mathbf{S}$) are promoted to the beginning of the grow order $\lambda_Y$. This is illustrated in Figure 5 for variable 7, which depicts the grow phase of two consecutively examined variables 5 and 7. In this figure, the curved edges show the tests that are inferred by $I_{\text{GSIMN}}$ during the grow phase of variable 5. The grow order of 7 changes from $\lambda_7 = [2, 6, 3, 0, 4, 1, 5]$ to $\lambda_7 = [3, 4, 6, 5, 2, 0, 1]$ after the grow phase of variable 5 is complete because the variables 5, 6, 4 and 3 were promoted (in that order) to the beginning of the queue. The rationale for this is the observation that this increases the number of tests inferred by GSIMN at the next step: The change in the examination and grow orders described above was chosen so that the inferred tests while learning the blanket of variable 7 match *exactly* those required by the algorithm in some future step. In particular, note that in the example the set of inferred dependencies between each variable found dependent with 5 before 7 are exactly those required during initial part of the grow phase of variable 7, shown highlighted in Figure 5(b) (the first four dependencies). These independence tests were inferred (not conducted), resulting in computational savings. In general, the last dependent variable of the grow phase of $X$ has the maximum number of dependences and independences inferred and this provides the rationale for its change in grow order and its selection by the algorithm to be examined next.

It can be shown that under the same assumptions as GSMN*, the structure returned by GSIMN is the correct one i.e., each set $\mathbf{B}^X$ computed by the GSIMN algorithm equals exactly the neighbors of $X$. The proof of correctness of GSIMN is based on correctness of GSMN* and is presented in Appendix B.

## 4.6 GSIMN Technical Implementation Details

In this section we discuss a number of practical issues that subtly influence the accuracy and efficiency of an implementation of GSIMN. One is the order of application of the I-SU, D-SU, I-Triangle and D-Triangle rules within the function $I_{\text{GSIMN}}$. Given an independence-query oracle, the order of application should not matter—assuming there are more than one rules for inferring the value of an independence, all of them are guaranteed to produce the same value due to the soundness of the axioms of Eqs. (1) (Pearl, 1988). In practice however, the oracle is implemented by statistical tests conducted on data which can be incorrect, as previously mentioned. Of particular importance is the observation that false independences are more likely to occur than false dependencies. One example of this is the case where the domain dependencies are weak—in this case any pair of variables connected (dependent) in the underlying true network structure may be incorrectly deemed independent if all paths between them are long enough. On the other hand, false dependencies are much more rare—the confidence threshold of $1-\alpha = 0.95$ of a statistical test tells us that the probability of a false dependence by chance alone is only 5%. Assuming i.i.d. data for each test, the chance of multiple false dependencies is even lower, decreasing exponentially fast. This practical observation i.e., that dependencies are typically more reliable than independences, provide the rationale for the way the $I_{\text{GSIMN}}$ algorithm works. In particular, $I_{\text{GSIMN}}$ prioritizes the application of rules whose antecedents contain dependencies first i.e., the D-Triangle and D-SU rules, followed by the I-Triangle and I-SU rules. In effect, this uses statistical results that are typically known with greater confidence before ones that are usually less reliable.

The second practical issue concerns efficient inference. The GSIMN algorithm uses a one-step inference procedure (shown in Algorithm 5) that utilizes a knowledge base $K = \{K_{XY}\}$ containing known independences and dependences for each pair of variables $X$ and $Y$. To implement this inference efficiently we utilize a data structure for $K$ for the purpose of storing and retrieving independence facts in constant time. It consists of two 2D arrays, one for dependencies and another for independencies. Each array is of $n \times n$ size, where $n$ is the number of variables in the domain. Each cell in this array corresponds to a pair of variables $(X, Y)$, and stores the known independences (dependences) between $X$ and $Y$ in the form of a list of conditioning sets. For each conditioning set $\mathbf{Z}$ in the list, the knowledge base $K_{XY}$ represents a known independence $(X \perp\!\!\!\perp Y \mid \mathbf{Z})$ (dependence $(X \not\perp\!\!\!\perp Y \mid \mathbf{Z})$). It is important to note that the length of each list is at most 2, as there are no more than two tests done between any variable $X$ and $Y$ during the execution of GSIMN (done during the growing and shrinking phases). Thus, it always takes a constant time to retrieve/store an independence (dependence), and therefore all inferences using the knowledge base are constant time as well. Also note that all uses of the Strong Union axiom by the $I_{\text{GSIMN}}$ algorithm are constant time as well, as they can be accomplished by testing the (at most two) sets stored in $K_{XY}$ for subset or superset inclusion.

## 5. Experimental Results

We evaluated the GSMN* and GSIMN algorithms on both artificial and real-world data sets. Through the experimental results presented below we show that the simple application of Pearl's inference rules in GSIMN algorithm results in a significant reduction in the number of tests performed when compared to GSMN* without adversely affecting the quality of the output network. In particular we report the following quantities:

- **Weighted number of tests.** The weighted number of tests is computed by the summation of the weight of each test executed, where the *weight* of test $(X, Y \mid \mathbf{Z})$ is defined as $2 + |\mathbf{Z}|$. This quantity reflects the time complexity of the algorithm (GSMN* or GSIMN) and can be used to assess the benefit in GSIMN of using inference instead of executing statistical tests on data. This is the standard method of comparison of independence-based algorithms and it is justified by the observation that the running time of a statistical test on triplet $(X, Y \mid \mathbf{Z})$ is proportional to the size $N$ of the data set and the number of variables involved in it i.e., $O(N(|\mathbf{Z}|+2))$ (and is not exponential in the number of variables involved as a naïve implementation might assume). This is because one can construct all non-zero entries in the contingency table used by the test by examining each data point in the data set exactly once, in time proportional to the number of variables involved in the test i.e., proportional to $|\{X, Y\} \cup \mathbf{Z}| = 2 + |\mathbf{Z}|$.

- **Execution time.** In order to assess the impact of inference in the running time (in addition to the impact of statistical tests), we report the execution time of the algorithm.

- **Quality of the resulting network.** We measure quality in two ways.

  - **Normalized Hamming distance.** The Hamming distance between the output network and the structure of the underlying model is another measure of the quality of the output network, when the actual network that was used to generate the data is known. The Hamming distance is defined as the number of "reversed" edges between these two network structures, i.e., the number of times an actual edge in the true network is missing in the returned network or an edge absent from the true network exists in the algorithm's output network. A value zero means that the output network has the correct structure. To be able to compare domains of different dimensionalities (number of variables $n$) we normalize it by $\binom{n}{2}$, the total number of node pairs in the corresponding domain.

  - **Accuracy.** For real-world data sets where the underlying network is unknown, no Hamming distance calculation is possible. In this case it is impossible to know the true value of any independence. We therefore approximate it by a statistical test on the entire data set, and use a limited, randomly chosen subset (1/3 of the data set) to learn the network. To measure accuracy we compare the result (`true` or `false`) of a number of conditional independence tests on the network output (using vertex separation), to the same tests performed on the full data set.

In all experiments involving data sets we used the $\chi^2$ statistical test for estimation of conditional independences. As mentioned above, rules of thumb exist that deem certain tests as potentially unreliable depending on the counts of the contingency table involved; for example, one such rule Cochran (1954) deems a test unreliable if more than 20% of the cells of the contingency table have less than 5 data points the test. Due to the requirement that an answer must be obtained by an independence algorithm conducting a test, we used the outcomes of such tests as well in our experiments. The effect of these possibly unreliable tests on the quality of the resulting network is measured by our accuracy measures, listed above.

In the next section we present results for domains in which the underlying probabilistic model is known. This is followed by real-world data experiments where no model structure is available.

## 5.1 Known-Model Experiments

In the first set of experiments the underlying model, called the *true model* or *true network*, is a known Markov network. The purpose of this set of experiments is to conduct a controlled evaluation of the quality of the output network through a systematic study of the algorithms' behavior under varying conditions of domain size (number of variables) and amount of dependencies (average node degree in the network).

Each true network that contains $n$ variables was generated randomly as follows: the network was initialized with $n$ nodes and no edges. A user-specified parameter of the network structure is the average node degree $\tau$ that equals the average number of neighbors per node. Given $\tau$, for every node its set of neighbors was determined randomly and uniformly by selecting the first $\tau \frac{n}{2}$ pairs in a random permutation of all possible pairs. The factor $1/2$ is necessary because each edge contributes to the degree of two nodes.

We conducted two types of experiments using known network structure: *Exact learning* experiments and *sample-based* experiments.

### 5.1.1 Exact Learning Experiments

In this set of known-model experiments, we assume that the result of all statistical queries asked by the GSMN* and GSIMN algorithms were available, which assumes the existence of an oracle that can answer independence queries. When the underlying model is known, this oracle can be implemented through vertex separation. The benefits of querying the true network for independence are two: First, it ensures faithfulness and correctness of the independence query results, which allows the evaluation of the algorithms under their assumptions for correctness. Second, these tests can be performed much faster than actual statistical tests on data. This allowed us to evaluate our algorithms in large networks—we were able to conduct experiments of domains containing up to 100 variables.

We first report the weighted number of tests executed by GSMN* with and without propagation and GSIMN. Our results are summarized in Figure 6, which shows the ratio between the weighted number of tests of GSIMN and the two versions of GSMN*. One

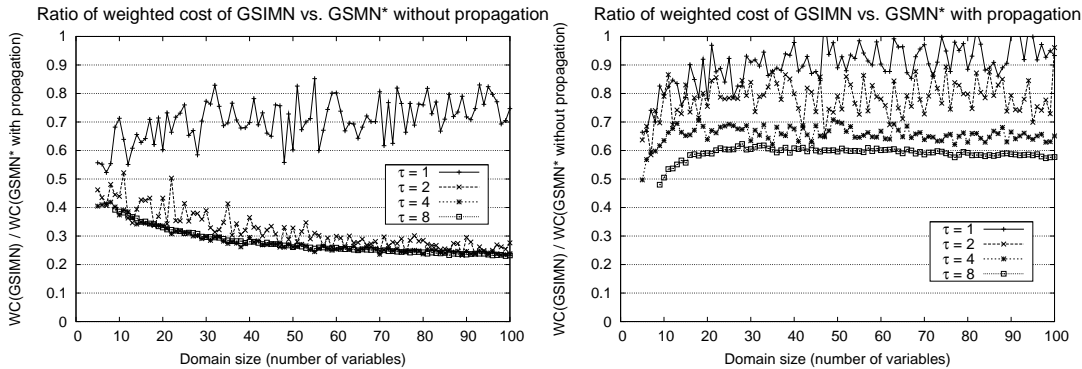

Figure 6: Ratio of the weighted number of tests of GSIMN over GSMN* without propagation (**left plot**) and with propagation (**right plot**) for network sizes (number of nodes) up to $n = 100$ of average degree $\tau = 1, 2, 4,$ and 8.

---

**Algorithm 6** $I_{\text{FCH}}(X, Y, \mathbf{S}, \mathbf{F}, \mathbf{T})$. Forward-chaining implementation of independence test $I_{\text{GSIMN}}(X, Y, \mathbf{S}, \mathbf{F}, \mathbf{T})$.

---
1: /* *Query knowledge base.* */
2: **if** $\exists (\mathbf{S}, t) \in K_{XY}$ **then**
3:     **return** $t$
4: $t \leftarrow$ result of test $(X, Y \mid \mathbf{S})$ /* $t = \texttt{true}$ *iff test* $(X, Y \mid \mathbf{S})$ *returns independence.* */
5: Add $(\mathbf{S}, t)$ to $K_{XY}$ and $K_{YX}$.
6: Run forward-chaining inference algorithm on $K$, update $K$.
7: **return** $t$

---

hundred true networks were generated randomly for each pair $(n, \tau)$, and the figure shows the mean value. We can see that the limiting reduction (as $n$ grows large) in weighted number of tests depends primarily on the average degree parameter $\tau$. The reduction of GSIMN for large $n$ and dense networks ($\tau = 8$) is approximately 40% compared to GSMN* with propagation and 75% compared to GSMN* without the propagation optimization, demonstrating the benefit of GSIMN vs. GSMN* in terms of number of tests executed.

One reasonable question about the performance of GSIMN is to what extent its inference procedure is *complete* i.e., from all those tests that GSIMN needs during its operation, how does the number of tests that it infers (by applying a single step of backward chaining on the Strong Union axiom and the Triangle theorem, rather than executing a statistical test on data) compare to the number of tests that *can* be inferred (for example using a complete automated theorem prover on Eqs. (1))? To measure this, we compared the number of tests done by GSIMN with the number done by an alternative algorithm, which we call GSIMN-FCH (GSIMN with Forward Chaining). GSIMN-FCH differs from GSIMN in function $I_{\text{FCH}}$, shown in Algorithm 6, which replaces function $I_{\text{GSIMN}}$ of GSIMN. $I_{\text{FCH}}$ exhaustively produces all independence statements that can be inferred through the properties of Eqs. (1) using a forward-chaining procedure. This process iteratively builds a knowledge base $K$ containing the truth value of conditional independence predicates. Whenever the outcome of a test is required, $K$ is queried (line 2 of $I_{\text{FCH}}$ in Algorithm 6). If the value of the test is found in $K$, it is returned (line 3). If not, GSIMN-FCH performs the test and uses the result in a standard forward-chaining automatic theorem prover subroutine (line 6) to produce all independence statements that can be inferred by the test result and $K$, adding these new facts to $K$.

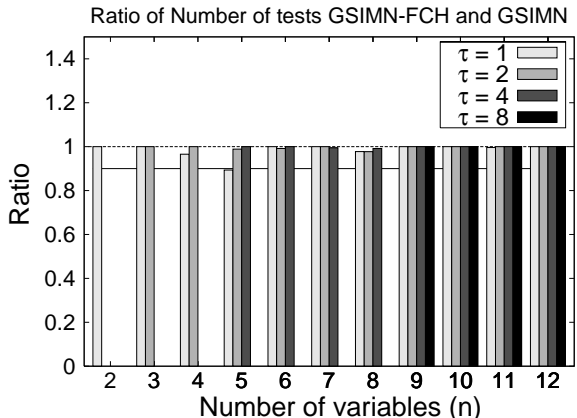

Figure 7: Ratio of number of tests of GSIMN-FCH over GSIMN for network sizes (number of variables) $n = 2$ to $n = 13$ and average degrees $\tau = 1$, 2, 4, and 8.

A comparison of the number of tests executed by GSIMN vs. GSIMN-FCH is presented in Figure 7, which shows the ratio of the number of tests of GSIMN over GSIMN-FCH. The figure shows the mean value over four runs, each corresponding to a network generated randomly for each pair $(n, \tau)$, for $\tau = 1$, 2, 4 and 8 and $n$ up to 12. Unfortunately, after two days of execution GSIMN-FCH was unable to complete execution on domains containing 13 variables or more. We therefore present results for domain sizes up to 12 only. The figure shows that for $n \geq 9$, and every $\tau$ the ratio is exactly 1 i.e., all tests inferable were produced by the use of the Triangle theorem in GSIMN. For smaller domains, the ratio is above 0.95 with the exception of a single case, $(n = 5, \tau = 1)$.

### 5.1.2 SAMPLE-BASED EXPERIMENTS

In this set of experiments we evaluate GSMN* (with and without propagation) and GSIMN on data sampled from the true model. This allows a more realistic assessment of the performance of our algorithms. The data were sampled from the true (known) Markov network using Gibbs sampling.

In the exact learning experiments of the previous section only the structure of the true network was required, generated randomly in the fashion described above. To sample data from a known structure however, one also needs to specify the network parameters. For each random network, the parameters determine the strength of dependencies among connected variables in the graph. Following Agresti (2002), we used the *log-odds ratio* as a measure of the strength of the probabilistic influence between two binary variables $X$ and $Y$, defined as

$$\theta_{XY} = \log \frac{\Pr(X = 0, Y = 0) \Pr(X = 1, Y = 1)}{\Pr(X = 0, Y = 1) \Pr(X = 1, Y = 0)}.$$

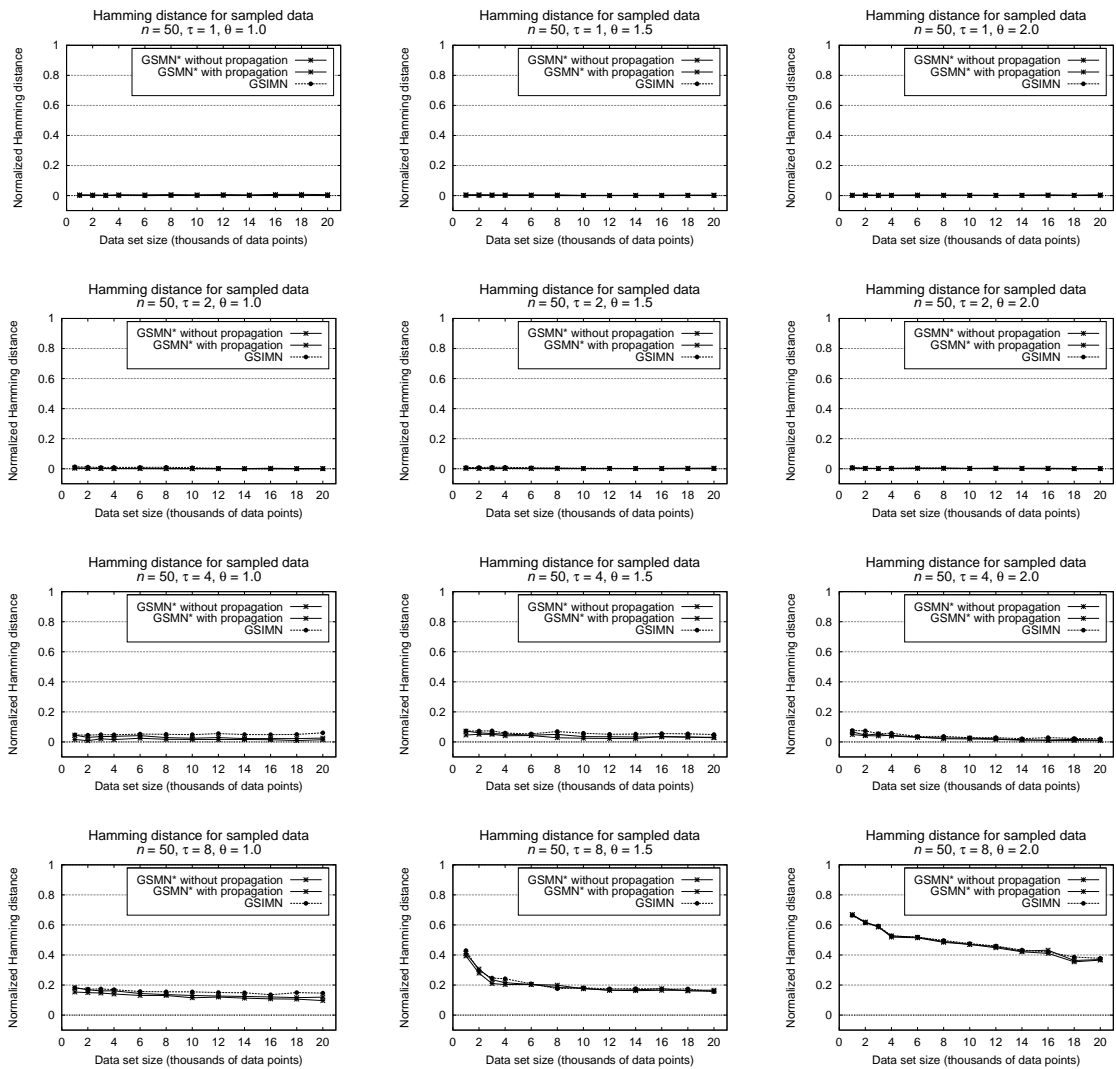

Figure 8: Normalized Hamming distances between the true network and the network output by GSMN* (with and without propagation) and GSIMN for domain size $n = 50$ and average degrees $\tau = 1, 2, 4, 8$.

The network parameters were generated randomly so that the log-odds ratio between every pair of variables connected by an edge in the graph has a specified value. In this set of experiments, we used values of $\theta = 1$, $\theta = 1.5$ and $\theta = 2$ for every such pair of variables in the network.

Figures 8 and 9 show plots of the normalized Hamming distance between the true network and that output by the GSMN* (with and without propagation) and GSIMN for domain sizes of $n = 50$ and $n = 75$ variables, respectively. These plots show that the Hamming distance of GSIMN is comparable to the ones of the GSMN* algorithms for both domain sizes $n = 50$ and $n = 75$, all average degrees $\tau = 1, 2, 4, 8$ and log-odds ratios $\theta = 1$, $\theta = 1.5$ and $\theta = 2$. This reinforces the claim that inference done by GSIMN has a small impact on the quality of the output networks.

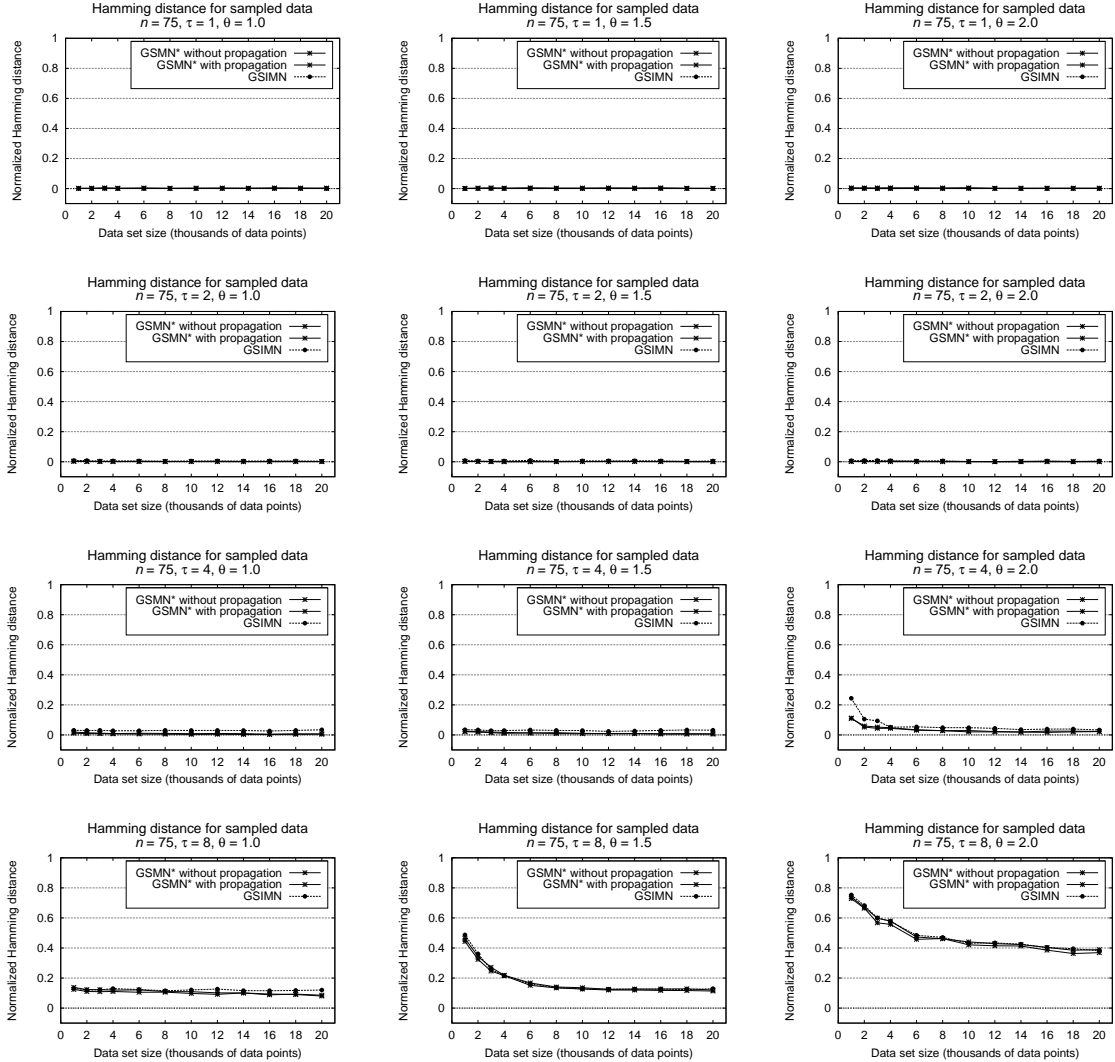

Figure 9: Normalized Hamming distance results as in Figure 8 but for domain size $n = 75$.

Figure 10 shows the weighted number of tests of GSIMN vs. GSMN* (with and without propagation) for a sampled data set of 20,000 points for domains $n = 50$, and $n = 75$, average degree parameters $\tau = 1, 2, 4$, and 8 and log-odds ratios $\theta = 1, 1.5$ and 2. GSIMN shows a reduced weighted number of tests with respect to GSMN* without propagation in all cases and compared to GSMN* with propagation in most cases (with the only exceptions of $(\tau = 4, \theta = 2)$ and $(\tau = 8, \theta = 1.5)$). For sparse networks and weak dependences i.e., $\tau = 1$, this reduction is larger than 50% for both domain sizes, a reduction much larger than the one observed for the exact learning experiments. The actual execution times for various data set sizes and network densities are shown in Figure 11 for the largest domain of $n = 75$, and $\theta = 1$, verifying the reduction in cost of GSIMN for various data set sizes. Note that the reduction is proportional to the number of data points; this is reasonable as each test executed must go over the entire data set once to construct the contingency table. This confirms our claim that the cost of inference of GSIMN is small (constant time per test, see discussion in Section 4.6) compared to the execution time of the tests themselves, and indicates increasing cost benefits of the use of GSIMN for even large data sets.

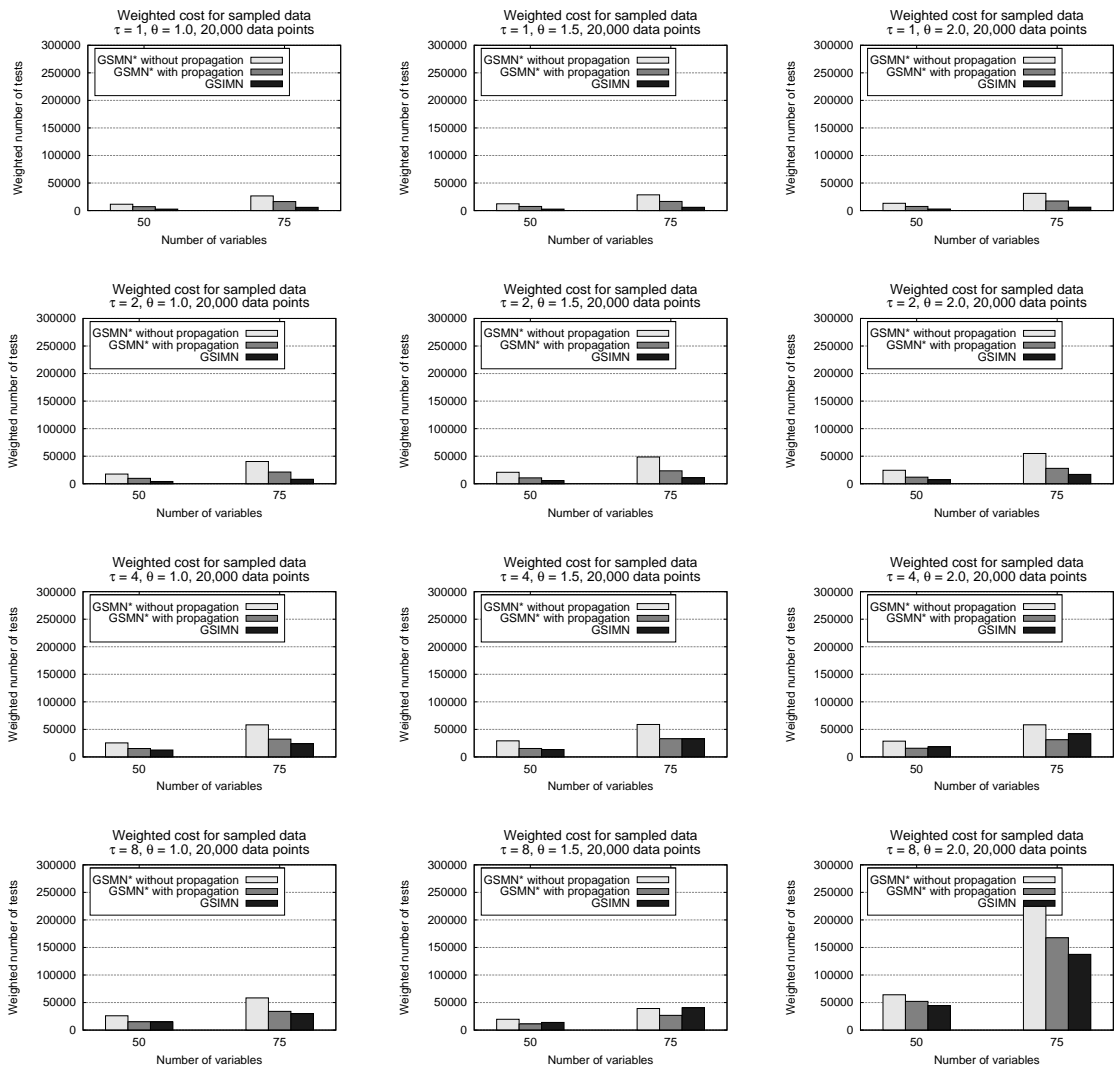

Figure 10: Weighted number of tests executed by GSMN* (with and without propagation) and GSIMN for $|D| = 20{,}000$, for domains sizes $n = 50$ and $75$, average degree parameters $\tau = 1$, $2$, $4$, and $8$, and log-odds ratios $\theta = 1$, $1, 5$, and $2$.

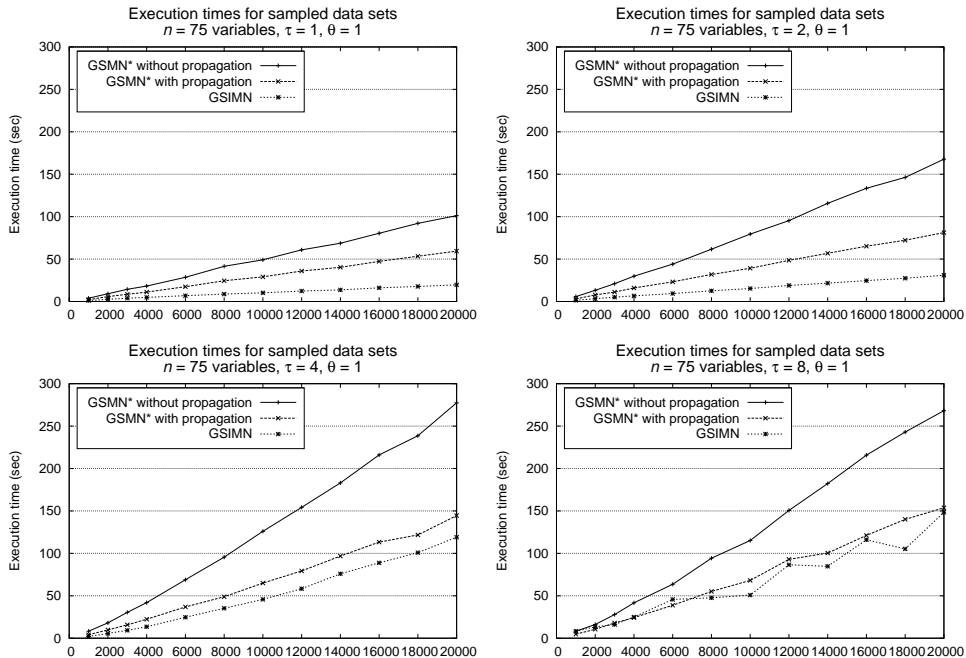

Figure 11: Execution times for sampled data experiments for $\theta = 1$, $\tau = 1, 2$ (**top row**) and $\tau = 4, 8$ (**bottom row**) for a domain of $n = 75$ variables.

### 5.1.3 Real-World Network Sampled Data Experiments

We also conducted sampled data experiments on well-known real-world networks. As there is no known repository of Markov networks drawn from real-world domains, we instead utilized well-known Bayesian networks that are widely used in Bayesian network research and are available from a number of repositories.[1] To generate Markov networks from these Bayesian network structures we used the process of *moralization* (Lauritzen, 1996) that consists of two steps: (a) connect each pair of nodes in the Bayesian network that have a common child with an undirected edge and (b) remove directions of all edges. This results in a Markov network in which the local Markov property is valid i.e., each node is conditionally independent of all other nodes in the domain given its direct neighbors. During this procedure some conditional independences may be lost. This, however, does not affect the accuracy results because we compare the independencies of the output network with those of the moralized Markov network (as opposed to the Bayesian network).

We conducted experiments using 5 real-world domains: Hailfinder, Insurance, Alarm, Mildew, and Water. For each domain we sampled a varying number of data points from its corresponding Bayesian network using logic sampling (Henrion, 1988), and used it as input to the GSMN* (with and without propagation) and GSIMN algorithms. We then compared the network output from each of these algorithms to the original moralized network using the normalized Hamming distance metric previously described. The results are shown in

---

1. We used `http://compbio.cs.huji.ac.il/Repository/`. Accessed on December 5, 2008.

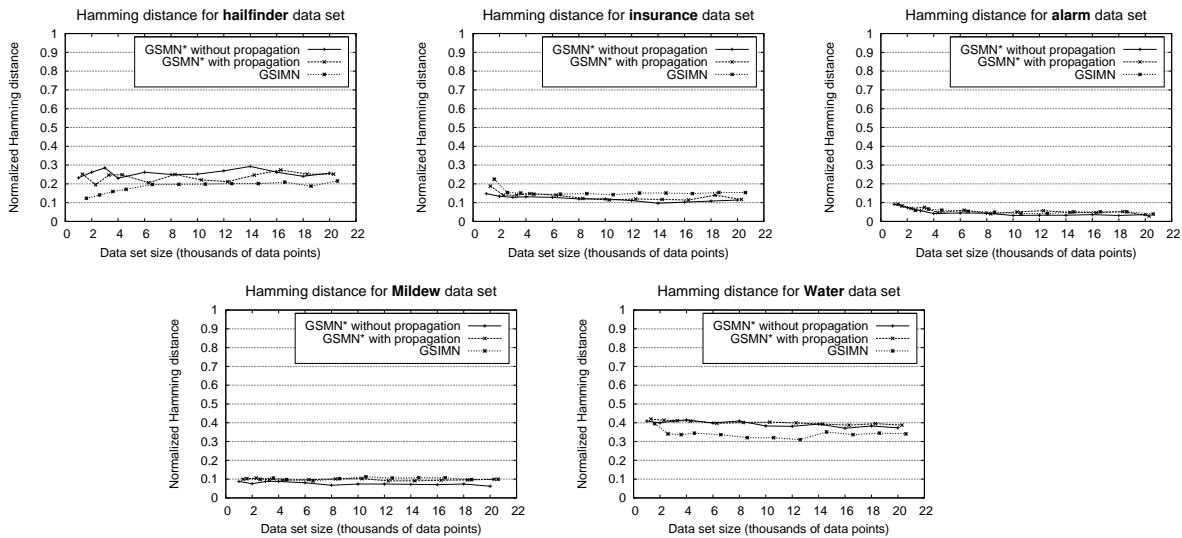

Figure 12: Normalized Hamming distance of the network output by GSMN* (with and without propagation) and GSIMN with the true Markov networks network using varying data set sizes sampled from Markov networks for various real-world domains modeled by Bayesian networks.

Fig. 12 and indicate that the distances produced from the three algorithms are similar. In some cases (e.g., Water and Hailfinder) the network resulting from the use of GSIMN is actually better (of smaller Hamming distance) than the ones output by the GSMN* algorithms.

We also measured the weighted cost of the three algorithms for each of these domains, shown in Fig. 13. The plots show a significant decrease in the weighted number of tests for GSIMN with respect to both GSMN* algorithms: the cost of GSIMN is 66% of the cost of GSMN* without propagation on average, a savings of 34%, while the cost of GSIMN is 28% of the cost of GSMN* without propagation on average, a savings of 72%.

## 5.2 Real-World Data Experiments

While the artificial data set studies of the previous section have the advantage of allowing a more controlled and systematic study of the performance of the algorithms, experiments on real-world data are necessary for a more realistic assessment of their performance. Real data are more challenging because they may come from non-random topologies (e.g., a possibly irregular lattice in many cases of spatial data) and the underlying probability distribution may not be faithful.

We conducted experiments on a number of data sets obtained from the UCI machine learning data set repository (Newman, Hettich, Blake, & Merz, 1998). Continuous variables in the data sets were discretized using a method widely recommended in introductory statistics texts (Scott, 1992); it dictates that the optimal number of equally-spaced discretization bins for each continuous variable is $k = 1 + \log_2 N$, where $N$ is the number of points in the data set. For each data set and each algorithm, we report the weighted number of conditional independence tests conducted to discover the network and the accuracy, as defined below.

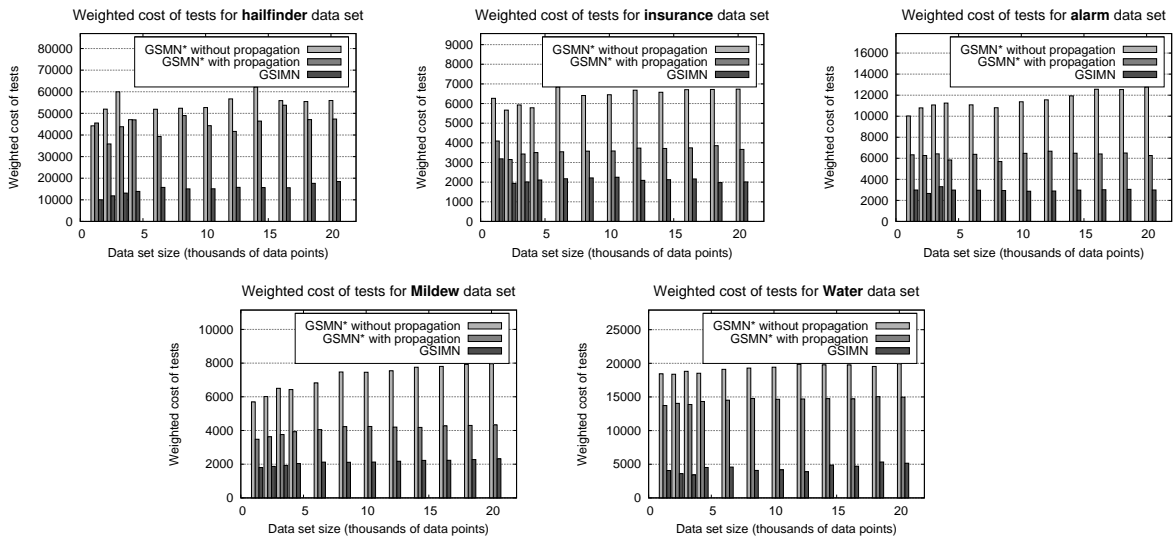

Figure 13: Weighted cost of tests conducted by the GSMN* (with and without propagation) and GSIMN algorithms for various real-world domains modeled by Bayesian networks.

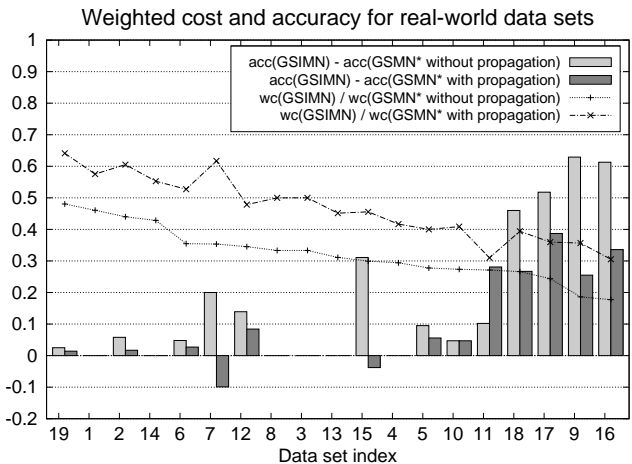

Figure 14: Ratio of the weighted number of tests of GSIMN versus GSMN* and difference between the accuracy of GSIMN and GSMN* on real data sets. Ratios smaller that 1 and positive bars indicate an advantage of GSIMN over GSMN*. The numbers in the x-axis are indices of the data sets as shown in Table 1.

Table 1: Weighted number of tests and accuracy for several real-world data sets. For each evaluation measure, the best performance between GSMN* (with and without propagation) and GSIMN is indicated in bold. The number of variables in the domain is denoted by $n$ and the number of data points in each data set by $N$.

| Data set | | | | Weighted number of tests | | | Accuracy | | |
|---|---|---|---|---|---|---|---|---|---|
| # | Name | $n$ | $N$ | GSMN* (w/o prop.) | GSMN* (w/ prop.) | GSIMN | GSMN* (w/o prop.) | GSMN* (w/ prop.) | GSIMN |
| 1 | echocardiogram | 14 | 61 | 1311 | 1050 | **604** | 0.244 | 0.244 | 0.244 |
| 2 | ecoli | 9 | 336 | 425 | 309 | **187** | 0.353 | 0.394 | **0.411** |
| 3 | lenses | 5 | 24 | 60 | 40 | **20** | 0.966 | 0.966 | 0.966 |
| 4 | hayes-roth | 6 | 132 | 102 | 72 | **30** | 0.852 | 0.852 | 0.852 |
| 5 | hepatitis | 20 | 80 | 1412 | 980 | **392** | 0.873 | 0.912 | **0.968** |
| 6 | cmc | 10 | 1473 | 434 | 292 | **154** | 0.746 | 0.767 | **0.794** |
| 7 | balance-scale | 5 | 625 | 82 | 47 | **29** | 0.498 | **0.797** | 0.698 |
| 8 | baloons | 5 | 20 | 60 | 40 | **20** | 0.932 | 0.932 | 0.932 |
| 9 | flag | 29 | 194 | 5335 | 2787 | **994** | 0.300 | 0.674 | **0.929** |
| 10 | tic-tac-toe | 10 | 958 | 435 | 291 | **119** | 0.657 | 0.657 | **0.704** |
| 11 | bridges | 12 | 70 | 520 | 455 | **141** | 0.814 | 0.635 | **0.916** |
| 12 | car | 7 | 1728 | 194 | 140 | **67** | 0.622 | 0.677 | **0.761** |
| 13 | monks-1 | 7 | 556 | 135 | 93 | **42** | 0.936 | 0.936 | 0.936 |
| 14 | haberman | 5 | 306 | 98 | 76 | **42** | 0.308 | 0.308 | 0.308 |
| 15 | nursery | 9 | 12960 | 411 | 270 | **123** | 0.444 | **0.793** | 0.755 |
| 16 | crx | 16 | 653 | 1719 | 999 | **305** | 0.279 | 0.556 | **0.892** |
| 17 | imports-85 | 25 | 193 | 4519 | 3064 | **1102** | 0.329 | 0.460 | **0.847** |
| 18 | dermatology | 35 | 358 | 9902 | 6687 | **2635** | 0.348 | 0.541 | **0.808** |
| 19 | adult | 10 | 32561 | 870 | 652 | **418** | 0.526 | 0.537 | **0.551** |

Because for real-world data the structure of the underlying Bayesian network (if any) is unknown, it is impossible to measure the Hamming distance of the resulting network structure. Instead, we measured the estimated accuracy of a network produced by GSMN* or GSIMN by comparing the result (`true` or `false`) of a number of conditional independence tests on the network learned by them (using vertex separation) to the result of the same tests performed on the data set (using a $\chi^2$ test). This approach is similar to estimating accuracy in a classification task over unseen instances but with inputs here being triplets $(X, Y, \mathbf{Z})$ and the class attribute being the value of the corresponding conditional independence test. We used 1/3 of the real-world data set (randomly sampled) as input to GSMN* and GSIMN and the entire data set to the $\chi^2$ test. This corresponds to the hypothetical scenario that a much smaller data set is available to the researcher, and approximates the true value of the test by its outcome on the entire data set. Since the number of possible tests is exponential, we estimated the independence accuracy by sampling 10,000 triplets $(X, Y, \mathbf{Z})$ randomly, evenly distributed among all possible conditioning set sizes $m \in \{0, \ldots, n - 2\}$ (i.e., $10000/(n - 1)$ tests for each $m$). Each of these triplets was constructed as follows: First, two variables $X$ and $Y$ were drawn randomly from $\mathbf{V}$. Second, the conditioning set was determined by picking the first $m$ variables from a random permutation of $\mathbf{V} - \{X, Y\}$. Denoting by $\mathcal{T}$ this set of 10,000 triplets, by $t \in \mathcal{T}$ a triplet, by $I_{\text{data}}(t)$ the result of a test performed on the entire data set and by $I_{\text{network}}(t)$ the result of a test performed on the network output by either GSMN* or GSIMN, the estimated accuracy is defined as:

$$\widehat{accuracy} = \frac{1}{|\mathcal{T}|}\left|\left\{t \in \mathcal{T} \mid I_{\text{network}}(t) = I_{\text{data}}(t)\right\}\right|.$$

For each of the data sets, Table 1 shows the detailed results for accuracy and the weighted number of tests for the GSMN* and GSIMN algorithms. These results are also plotted in Figure 14, with the horizontal axis indicating the data set index appearing in the first column of Table 1. Figure 14 plots two quantities in the same graph for these real-world data sets: the ratio of the weighted number of tests of GSIMN versus the two GSMN* algorithms and the difference of their accuracies. For each data set, an improvement of GSIMN over GSMN* corresponds to a number smaller than 1 for the ratios and a positive histogram bar for the accuracy differences. We can observe that GSIMN reduced the weighted number of tests on every data set, with maximum savings of 82% over GSMN* without propagation (for the "crx" data set) and 60% over GSMN* with propagation (for the "crx" data set as well). Moreover, in 11 out of 19 data sets GSIMN resulted in improved accuracy, 6 in a tie and only 2 in somewhat reduced accuracy compared to GSMN* with propagation (for the "nursery" and "balance-scale" data sets).

## 6. Conclusions and Future Research

In this paper we presented two algorithms, GSMN* and GSIMN, for learning efficiently the structure of a Markov network of a domain from data using the independence-based approach (as opposed to NP-hard algorithms based on maximum likelihood estimation) We evaluated their performance through measurement of the weighted number of tests they require to learn the structure of the network and the quality of the networks learned from both artificial and real-world data sets. GSIMN showed a decrease in the vast majority of artificial and real-world domains in an output network quality comparable to that of GSMN*, with some cases showing improvement. In addition, GSIMN was shown to be nearly optimal in the number of tests executed compared to GSIMN-FCH, which uses an exhaustive search to produce all independence information that can inferred from Pearl's axioms. Some directions of future research include an investigation into the way the topology of the underlying Markov network affects the number of tests required and quality of the resulting network, especially for commonly occurring topologies such as grids. Another research topic is the impact on number of tests of other examination and grow orderings of the variables.

## Acknowledgments

We thank Adrian Silvescu for insightful comments on accuracy measures and general advice on the theory of undirected graphical models.

## Appendix A. Correctness of GSMN*

For each variable $X \in \mathbf{V}$ examined during the main loop of the GSMN* algorithm (lines 10–39), the set $\mathbf{B}^X$ of variable $X \in \mathbf{V}$ is constructed by growing and shrinking a set $\mathbf{S}$, starting from the empty set. $X$ is then connected to each member of $\mathbf{B}^X$ to produce the structure of a Markov network. We prove that this procedure returns the actual Markov network structure of the domain.

For the proof of correctness we make the following assumptions.

- The axioms of Eqs. (1) hold.

- The probability distribution of the domain is strictly positive (required for Intersection axiom to hold).

- Tests are conducted by querying an oracle, which returns its true value in the underlying model.

The algorithm examines every variable $Y \in \lambda^X$ for inclusion to $\mathbf{S}$ (and thus to $\mathbf{B}^X$) during the grow phase (lines 18 to 33) and, if $Y$ was added to $\mathbf{S}$ during the grow phase, it considers it for removal during the shrinking phase (lines 34 to 37). Note that there is only one test executed between $X$ and $Y$ during the growing phase of $X$; we call this the *grow test* of $Y$ on $X$ (line 23). Similarly, there is one or no tests executed between $X$ and $Y$ during the shrinking phase; this test (if executed) is called the *shrink test* of $Y$ on $X$ (line 36).

The general idea behind the proof is to show that, while learning the blanket of $X$, variable $Y$ is in $\mathbf{S}$ by the end of the shrinking phase if and only if the dependence $(X \not\perp\!\!\!\perp Y \mid \mathbf{V} - \{X, Y\})$ between $X$ and $Y$ holds (which, according to Theorem 2 at the end of the Appendix, implies there is an edge between $X$ and $Y$). We can immediately prove one direction.

**Lemma 1.** *If $Y \notin \mathbf{S}$ at the end of the shrink phase, then $(X \perp\!\!\!\perp Y \mid \mathbf{V} - \{X, Y\})$.*

*Proof.* Let us assume that $Y \notin \mathbf{S}$ by the end of the shrink phase. Then, either $Y$ was not added to set $\mathbf{S}$ during the grow phase (i.e., line 24 was never reached), or removed from it during the shrink phase (i.e., line 37 was reached). The former can only be true if $(p_{XY} > \alpha)$ in line 22 (indicating $X$ and $Y$ are unconditionally independent) or $Y$ was found independent of $X$ in line 23. The latter can only be true if $Y$ was found independent of $X$ in line 36. In all cases then $\exists \mathbf{A} \subseteq \mathbf{V} - \{X, Y\}$ such that $(X \perp\!\!\!\perp Y \mid \mathbf{A})$, and by Strong Union then $(X \perp\!\!\!\perp Y \mid \mathbf{V} - \{X, Y\})$. $\qquad\square$

The opposite direction is proved in Lemma 6 below. However, its proof is more involved, requiring a few auxiliary lemmas, observations, and definitions. The two main auxiliary Lemmas are 4 and 5. Both use the lemma presented next (Lemma 2) inductively to extend the conditioning set of dependencies found by the grow and shrink tests between $X$ and $Y$, to all the remaining variables $\mathbf{V} - \{X, Y\}$. The Lemma shows that, if a certain independence holds, the conditioning set of a dependence can be increased by one variable.

**Lemma 2.** *Let $X, Y \in \mathbf{V}$, $\mathbf{Z} \subseteq \mathbf{V} - \{X, Y\}$, and $\mathbf{Z}' \subseteq \mathbf{Z}$. Then $\forall W \in \mathbf{V}$,*

$$(X \not\perp\!\!\!\perp Y \mid \mathbf{Z}) \ \text{and} \ (X \perp\!\!\!\perp W \mid \mathbf{Z}' \cup \{Y\}) \implies (X \not\perp\!\!\!\perp Y \mid \mathbf{Z} \cup \{W\}).$$

*Proof.* We prove by contradiction, and make use of the axioms of *Intersection* (I), *Strong Union* (SU), and *Decomposition* (D). Let us assume that $(X \not\perp\!\!\!\perp Y \mid \mathbf{Z})$ and $(X \perp\!\!\!\perp W \mid \mathbf{Z}' \cup \{Y\})$ but $(X \perp\!\!\!\perp Y \mid \mathbf{Z} \cup \{W\})$. Then

$$(X \perp\!\!\!\perp Y \mid \mathbf{Z} \cup \{W\}) \text{ and } (X \perp\!\!\!\perp W \mid \mathbf{Z}' \cup \{Y\})$$
$$\overset{\text{SU}}{\Longrightarrow} \quad (X \perp\!\!\!\perp Y \mid \mathbf{Z} \cup \{W\}) \text{ and } (X \perp\!\!\!\perp W \mid \mathbf{Z} \cup \{Y\})$$
$$\overset{\text{I}}{\Longrightarrow} \quad (X \perp\!\!\!\perp \{Y, W\} \mid \mathbf{Z})$$
$$\overset{\text{D}}{\Longrightarrow} \quad (X \perp\!\!\!\perp Y \mid \mathbf{Z}) \wedge (X \perp\!\!\!\perp W \mid \mathbf{Z})$$
$$\Longrightarrow \quad (X \perp\!\!\!\perp Y \mid \mathbf{Z}).$$

This contradicts the assumption $(X \not\perp\!\!\!\perp Y \mid \mathbf{Z})$. $\qquad\square$

We now introduce notation and definitions and prove auxiliary lemmas.

We denote by $\mathbf{S}_G$ the value of $\mathbf{S}$ at the end of the grow phase (line 34) i.e., the set of variables found dependent of $X$ during the grow phase, and by $\mathbf{S}_S$ the value of $\mathbf{S}$ at the end of the shrink phase (line 39). We also denote by $\mathbf{G}$ the set of variables found independent of $X$ during the grow phase and by $\mathbf{U} = [U_0, \ldots, U_k]$ the sequence of variables shrunk from $\mathbf{B}^X$, i.e., found independent of $X$ during the shrink phase. The sequence $\mathbf{U}$ is assumed ordered as follows: if $i < j$ then variable $U_i$ was found independent from $X$ before $U_j$ during the shrinking phase. A prefix of the first $i$ variables $[U_0, \ldots, U_{i-1}]$ of $\mathbf{U}$ is denoted by $\mathbf{U}_i$. For some test $t$ performed during the algorithm, we define $k(t)$ as the integer such that $\mathbf{U}_{k(t)}$ is the prefix of $\mathbf{U}$ containing the variables that were found independent of $X$ in this loop before $t$. Furthermore, we abbreviate $\mathbf{U}_{k(t)}$ by $\mathbf{U}_t$.

From the definition of $\mathbf{U}$ and the fact that in the grow phase the conditioning set increases by dependent variables only, we can immediately make the following observation:

**Observation 1.** *For some variable $U_i \in \mathbf{U}$, if $t$ denotes the shrink test performed between $X$ and $U_i$ then $\mathbf{U}_t = \mathbf{U}_{i-1}$.*

We can then relate the conditioning set of the shrink test $t$ with $\mathbf{U}_t$ as follows:

**Lemma 3.** *If $Y \in \mathbf{S}_S$ and $t = (X, Y \mid \mathbf{Z})$ is the shrink test of $Y$, then $\mathbf{Z} = \mathbf{S}_G - \mathbf{U}_t - \{Y\}$.*

*Proof.* According to line 36 of the algorithm, $\mathbf{Z} = \mathbf{S} - \{Y\}$. At the beginning of the shrink phase (line 34) $\mathbf{S} = \mathbf{S}_G$, but variables found independent afterward and until $t$ is conducted are removed from $\mathbf{S}$ in line 37. Thus, by the time $t$ is performed, $\mathbf{S} = \mathbf{S}_G - \mathbf{U}_t$ and the conditioning set becomes $\mathbf{S}_G - \mathbf{U}_t - \{Y\}$. $\qquad\square$

**Corollary 1.** $(X \perp\!\!\!\perp U_i \mid \mathbf{S}_G - \mathbf{U}_i)$.

*Proof.* The proof follows immediately from Lemma 3, Observation 1, and the fact that $\mathbf{U}_i = \mathbf{U}_{i-1} \cup \{U_i\}$. $\qquad\square$

The following two lemmas use Lemma 2 inductively to extend the conditioning set of the dependence between $X$ and a variable $Y$ in $\mathbf{S}_S$. The first lemma starts from the shrink test between $X$ and $Y$ (a dependence), and extends its conditioning set from $\mathbf{S}_S - \{Y\}$ (or equivalently $\mathbf{S}_G - \{Y\} - \mathbf{U}_t$ according to Lemma 3) to $\mathbf{S}_G - \{Y\}$.

**Lemma 4.** *If $Y \in \mathbf{S}_S$ and $t$ is the shrink test of $Y$, then $(X \not\!\perp\!\!\!\perp Y \mid \mathbf{S}_G - \{Y\})$.*

*Proof.* The proof proceeds by proving

$$(X \not\!\perp\!\!\!\perp Y \mid \mathbf{S}_G - \{Y\} - \mathbf{U}_i)$$

by induction on decreasing values of $i$, for $i \in \{0, 1, \ldots, k(t)\}$, starting at $i = k(t)$. The lemma then follows for $i = 0$ by noticing that $\mathbf{U}_0 = \varnothing$.

- **Base case** ($i = k(t)$): From Lemma 3, $t = (X, Y \mid \mathbf{S}_G - \{Y\} - \mathbf{U}_t)$, which equals $(X, Y \mid \mathbf{S}_G - \{Y\} - \mathbf{U}_{k(t)})$ by definition of $\mathbf{U}_t$. Since $Y \in \mathbf{S}_S$, it must be the case that $t$ was found dependent, i.e., $(X \not\!\perp\!\!\!\perp Y \mid \mathbf{S}_G - \{Y\} - \mathbf{U}_{k(t)})$.

- **Inductive step**: Let us assume that the statement is true for $i = m, 0 < m \leq k(t) - 1$:

$$(X \not\!\perp\!\!\!\perp Y \mid \mathbf{S}_G - \{Y\} - \mathbf{U}_m). \tag{2}$$

We need to prove that this is also true for $i = m - 1$:

$$(X \not\!\perp\!\!\!\perp Y \mid \mathbf{S}_G - \{Y\} - \mathbf{U}_{m-1}).$$

By Corollary 1, we have

$$(X \perp\!\!\!\perp U_m \mid \mathbf{S}_G - \mathbf{U}_m)$$

and by Strong Union,

$$(X \perp\!\!\!\perp U_m \mid (\mathbf{S}_G - \mathbf{U}_m) \cup \{Y\})$$

or

$$(X \perp\!\!\!\perp U_m \mid (\mathbf{S}_G - \mathbf{U}_m - \{Y\}) \cup \{Y\}). \tag{3}$$

From Eqs. (2), (3) and Lemma 2 we get the desired relation:

$$(X \not\!\perp\!\!\!\perp Y \mid (\mathbf{S}_G - \{Y\} - \mathbf{U}_m) \cup \{U_m\}) = (X \not\!\perp\!\!\!\perp Y \mid \mathbf{S}_G - \{Y\} - \mathbf{U}_{m-1}).$$

$\square$

**Observation 2.** *By definition of $\mathbf{S}_G$, we have that for every test $t = (X, Y \mid \mathbf{Z})$ performed during the grow phase, $\mathbf{Z} \subseteq \mathbf{S}_G$.*

The following lemma completes the extension of the conditioning set of the dependence between $X$ and $Y \in \mathbf{S}_S$ into the universe of variables $\mathbf{V} - \{X, Y\}$, starting from $\mathbf{S}_G - \{Y\}$ (where Lemma 4 left off) and extending it to $\mathbf{S}_G \cup \mathbf{G} - \{Y\}$.

**Lemma 5.** *If $Y \in \mathbf{S}_S$, then $(X \not\!\perp\!\!\!\perp Y \mid \mathbf{S}_G \cup \mathbf{G} - \{Y\})$.*

*Proof.* The proof proceeds by proving

$$(X \not\!\perp\!\!\!\perp Y \mid \mathbf{S}_G \cup \mathbf{G}_i - \{Y\})$$

by induction on increasing values of $i$ from 0 to $|\mathbf{G}|$, where $\mathbf{G}_i$ denotes the first $i$ elements of an arbitrary ordering of set $\mathbf{G}$.

- **Base Case** ($i = 0$): Follows directly from Lemma 4 for $i = 0$, since $\mathbf{G}_0 = \varnothing$.

- **Inductive Step**: Let us assume that the statement is true for $i = m, 0 \leq m < |\mathbf{G}|$:

$$(X \not\perp\!\!\!\perp Y \mid \mathbf{S}_G \cup \mathbf{G}_m - \{Y\}). \tag{4}$$

We need to prove that it is also true for $i = m + 1$:

$$(X \not\perp\!\!\!\perp Y \mid \mathbf{S}_G \cup \mathbf{G}_{m+1} - \{Y\}). \tag{5}$$

From Observation 2 the grow test of $G_m$ results in the independence:

$$(X \perp\!\!\!\perp G_m \mid \mathbf{Z}), \text{ where } \mathbf{Z} \subseteq \mathbf{S}_G.$$

By the Strong Union axiom this can become:

$$(X \perp\!\!\!\perp G_m \mid \mathbf{Z} \cup \{Y\}), \text{ where } \mathbf{Z} \subseteq \mathbf{S}_G \tag{6}$$

or equivalently

$$(X \perp\!\!\!\perp G_m \mid (\mathbf{Z} - \{Y\}) \cup \{Y\}), \text{ where } \mathbf{Z} \subseteq \mathbf{S}_G. \tag{7}$$

Since $\mathbf{Z} \subseteq \mathbf{S}_G \subseteq \mathbf{S}_G \cup \mathbf{G}_m$, we have that $\mathbf{Z} - \{Y\} \subseteq \mathbf{S}_G \cup \mathbf{G}_m$, and so from Eq. (4) and Lemma 2 we get the desired relation:

$$(X \not\perp\!\!\!\perp Y \mid (\mathbf{S}_G \cup \mathbf{G}_m - \{Y\}) \cup G_m) = (X \not\perp\!\!\!\perp Y \mid \mathbf{S}_G \cup \mathbf{G}_{m+1} - \{Y\}).$$

$$\square$$

Finally, we can prove that $X$ is dependent with every variable $Y \in \mathbf{S}_S$ given the universe $\mathbf{V} - \{X, Y\}$.

**Lemma 6.** *If* $Y \in \mathbf{S}_S$, *then* $(X \not\perp\!\!\!\perp Y \mid \mathbf{V} - \{X, Y\})$.

*Proof.* From Lemma 5,

$$(X \not\perp\!\!\!\perp Y \mid \mathbf{S}_G \cup \mathbf{G} - \{Y\})$$

It suffices then to prove that $\mathbf{S}_G \cup \mathbf{G} - \{Y\} = \mathbf{V} - \{X, Y\}$. In loop 6–9 of GSMN*, the queue $\lambda^X$ is populated with all elements in $\mathbf{V} - \{X\}$, and then, in line 21, $Y$ is removed from $\lambda^X$. The grow phase then partitions $\lambda^X$ into variables dependent of $X$ (set $\mathbf{S}_G$) and independent of $X$ (set $\mathbf{G}$). $\square$

**Corollary 2.** $Y \in \mathbf{S}_S \iff (X \not\perp\!\!\!\perp Y \mid \mathbf{V} - \{X, Y\})$.

*Proof.* Follows directly from Lemmas 1 and 6. $\square$

From the above Corollary we can now immediately show that the graph returned by connecting $X$ to each member of $\mathbf{B}^X = \mathbf{S}_S$ is exactly the Markov network of the domain using the following theorem, first published by Pearl and Paz (1985).

**Theorem 2.** *(Pearl & Paz, 1985) Every dependence model $M$ satisfying symmetry, decomposition, and intersection (Eqs. (1)) has a unique Markov network $G = (\mathbf{V}, \mathbf{E})$ produced by deleting from the complete graph every edge $(X, Y)$ for which $(X \perp\!\!\!\perp Y \mid \mathbf{V} - \{X, Y\})$ holds in $M$, i.e.,*

$$(X, Y) \notin \mathbf{E} \iff (X \perp\!\!\!\perp Y \mid \mathbf{V} - \{X, Y\}) \text{ in } M.$$

## Appendix B. Correctness of GSIMN

The GSIMN algorithm differs from GSMN* only by the use of test subroutine $I_{\text{GSIMN}}$ instead of $I_{\text{GSMN}^*}$ (Algorithms 5 and 4, respectively), which in turn differs by a number of additional inferences conducted to obtain the independencies (lines 8 to 22). These inferences are direct applications of the Strong Union axiom (which holds by assumption) and the Triangle theorem (which was proven to hold in Theorem 1). Using the correctness of GSMN* (proven in Appendix A) we can therefore conclude that the GSIMN algorithm is correct.